\documentclass[letterpaper]{article} 
\usepackage{aaai25}  
\usepackage{times}  
\usepackage{helvet}  
\usepackage{courier}  
\usepackage[hyphens]{url}  
\usepackage{graphicx} 
\usepackage{natbib}  
\usepackage{caption} 
\usepackage{algorithm}
\usepackage{algorithmic}

\usepackage{subcaption}
\usepackage{amsmath}
\usepackage{amsfonts}
\usepackage{xargs}
\usepackage{bm}
\usepackage{booktabs}
\usepackage{multirow}
\usepackage{amssymb}
\newcommand{\ud}{\mathrm{d}}

\newcommand{\bfa}{\mathbf{a}}
\newcommand{\bfb}{\mathbf{b}}

\newcommand{\bfd}{\mathbf{d}}

\newcommand{\bfh}{\mathbf{h}}

\newcommand{\bfk}{\mathbf{k}}

\newcommand{\bfq}{\mathbf{q}}

\newcommand{\bfs}{\mathbf{s}}

\newcommand{\bfv}{\mathbf{v}}

\newcommand{\bfz}{\mathbf{z}}

\newcommand{\bfA}{\mathbf{A}}

\newcommand{\bfC}{\mathbf{C}}

\newcommand{\bfK}{\mathbf{K}}

\newcommand{\bfV}{\mathbf{V}}
\newcommand{\bfW}{\mathbf{W}}

\newcommand{\rmq}{\mathrm{q}}

\newcommand{\rmP}{\mathrm{P}}

\newcommand{\rmS}{\mathrm{S}}

\newcommand{\calB}{\mathcal{B}}
\newcommand{\calC}{\mathcal{C}}

\newcommand{\calE}{\mathcal{E}}

\newcommand{\calG}{\mathcal{G}}
\newcommand{\calH}{\mathcal{H}}

\newcommand{\calK}{\mathcal{K}}
\newcommand{\calL}{\mathcal{L}}

\newcommand{\calN}{\mathcal{N}}
\newcommand{\calO}{\mathcal{O}}
\newcommand{\calP}{\mathcal{P}}

\newcommand{\calV}{\mathcal{V}}

\newcommand{\bbI}{\mathbb{I}}

\newcommand{\bbR}{\mathbb{R}}

\newcommand{\bmpsi}{\bm{\psi}}

\newcommand{\bmtheta}{\bm{\theta}}

\newcommand{\bmkappa}{\bm{\kappa}}

\newcommandx{\norm}[3][2={}, 3={}]{\vert\vert #1 \vert\vert_{#2}^{#3}}

\usepackage{xcolor}

\newcommandx{\sumlim}[3][1={i}, 2={1}]{\sum_{#1 = #2}^{#3}}
\newcommandx{\prodlim}[3][1={i}, 2={1}]{\prod_{#1 = #2}^{#3}}

\DeclareMathOperator*{\argmax}{arg\,max}

\newcommand{\MLP}{\mathrm{MLP}}
\usepackage{newfloat}
\usepackage{listings}
\DeclareCaptionStyle{ruled}{labelfont=normalfont,labelsep=colon,strut=off} 
\floatstyle{ruled}
\newfloat{listing}{tb}{lst}{}
\floatname{listing}{Listing}
\title{Neural Temporal Point Processes for Forecasting Directional Relations \\ in Evolving Hypergraphs}

\author {
    Tony Gracious,
    Arman Gupta,
    Ambedkar Dukkipati
}
\affiliations {
    Department of Computer Science and Automation\\
    Indian Institute of Science, Bangalore - 560012, INDIA \\
    \{tonygracious, arman, ambedkar\}@iisc.ac.in
}

\usepackage{bibentry}

\begin{document}

\maketitle

\begin{abstract}
   Forecasting relations between entities is paramount in the current era of data and AI. However, it is often overlooked that real-world relationships are inherently directional, involve more than two entities, and can change with time. 
   In this paper, we provide a comprehensive solution to the problem of forecasting directional relations in a general setting, where relations are higher-order, i.e., directed hyperedges in a hypergraph. This problem has not been previously explored in the existing literature. The primary challenge in solving this problem is that the number of possible hyperedges is exponential in the number of nodes at each event time. To overcome this, we propose a sequential generative approach that segments the forecasting process into multiple stages, each contingent upon the preceding stages, thereby reducing the search space involved in predictions of hyperedges. The first stage involves a temporal point process-based node event forecasting module that identifies the subset of nodes involved in an event. The second stage is a candidate generation module that predicts hyperedge sizes and adjacency vectors for nodes observing events. The final stage is a directed hyperedge predictor that identifies the truth by searching over the set of candidate hyperedges. To validate the effectiveness of our model, we compiled five datasets and conducted an extensive empirical study to assess each downstream task. Our proposed method achieves a performance gain of 32\% and 41\%  compared to the state-of-the-art pairwise and hyperedge event forecasting models, respectively, for the event type prediction. 
   
\end{abstract}

%


\section{Introduction}
The formation and evolution of real-world networks result from intricate relationships among entities. These relations can be characterized as (i) higher-order, (ii) directional, and (iii) temporal. For example, interactions via email involve a variable number of users engaged in directed relations that evolve with time. While there is a substantial body of literature focused on representation learning of networks, each study typically addresses these aspects in isolation. It is extremely challenging to study these networks in their most general form by considering all the above characteristics. This paper provides a comprehensive solution to this problem. 

Temporal network representations entail 
learning a model that aggregates historical data into finite-dimensional vectors on each node or entity. These vectors can subsequently be utilized to predict future interactions among the entities. 
The previous research has characterized relationships as pairwise edge formations within a network, employing temporal graph neural network-based models to derive representations~\citep{ShubhamEtAL:2019:AGenerativeModelForDynamicNetworksWithApplications,TrivediEtAL:2019:DyRepLearningRepresentationsoverDynamicGraphs,XuEtAL:2020:InductiveRepresentationLearningOnTemporalGraphs,LiuEtAL:2021:NeuralHigherOrderPatternMotifPreidctionInTemporalNetworks,CaoEtAL:2021:DeepStructuralPointProcessForLearningTemporalInteractionNetworks,RossiEtAL:2020:TemporalGraphNetworksForDeepLearningOnDynamicGraphs}. However, this setting has very limited applicability as most real-world relations, as mentioned earlier, are higher-order with groups of entities involved in each relation and sizes of the groups can vary in a network. These problems have been modeled and studied as hypergraphs in a static case~\cite{GhoshdastidarEtAL:2017:ConsistencyOfSpectralHypergraphPartitioningUnderPlantedPartitionModel,GhoshdastidarEtAL:2017:UniformHypergraphPartitioning}. While~\cite{GraciousEtAL:2023:DynamicRepresentationLearningWithTemporalPointProcessesForHigherOrderInteractionForecasting} is the first work that studies evolving hypergraphs, in this paper we study this problem in a very general setting considering that most real-world relations can be directional involving groups of interacting entities~\citep{KimEtAL:2022:ReciprocityInDirectedHypergraphsMeasuresFindingsAndGenerators}.  Such examples can be seen in sports such as football or cricket, where two groups compete. Similarly, in citation networks, authors draw upon the works of other authors to support their arguments. Hence, in this work, we model models relations as edge formation in a directed hypergraph. 

\begin{figure*}
  \centering
  \begin{subfigure}{0.5\textwidth}
    \includegraphics[width=\textwidth]{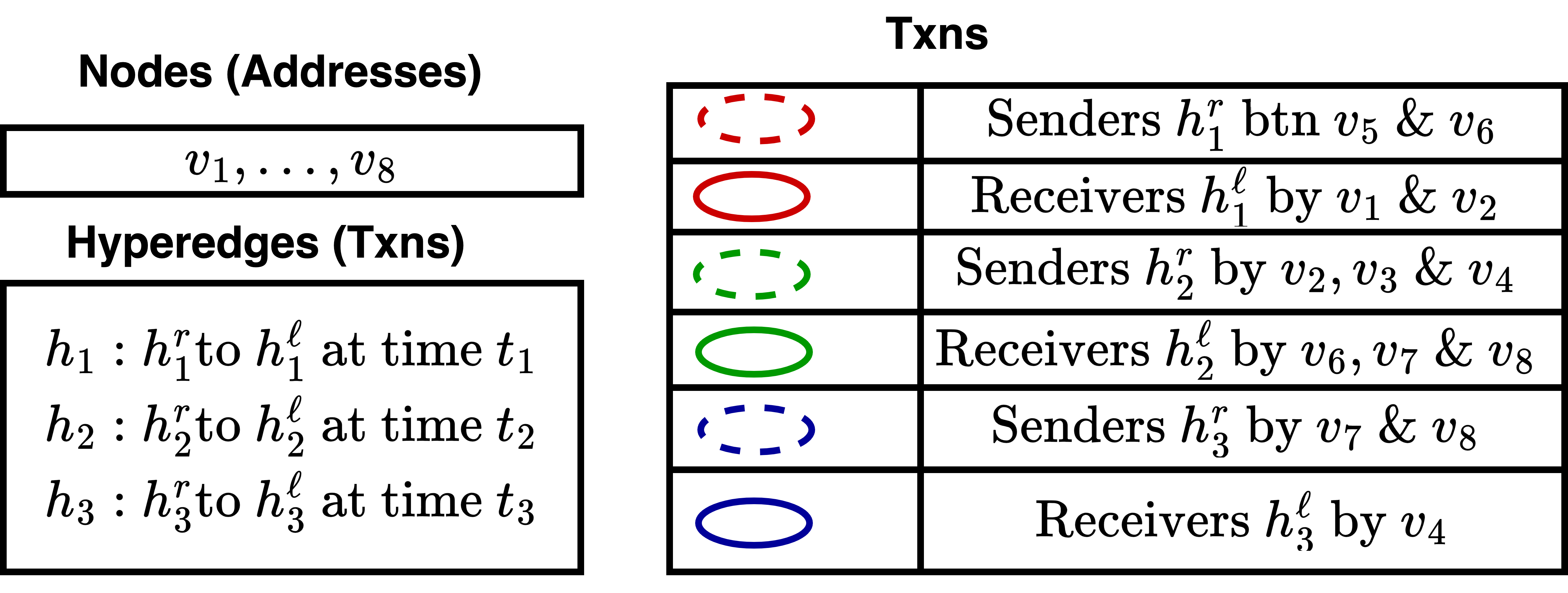}
    \caption{Transactions}
    \label{fig:temporal_txn_dataset}
  \end{subfigure}
  \hspace{1cm}
  \begin{subfigure}{0.25\textwidth}
    \includegraphics[width=\textwidth]{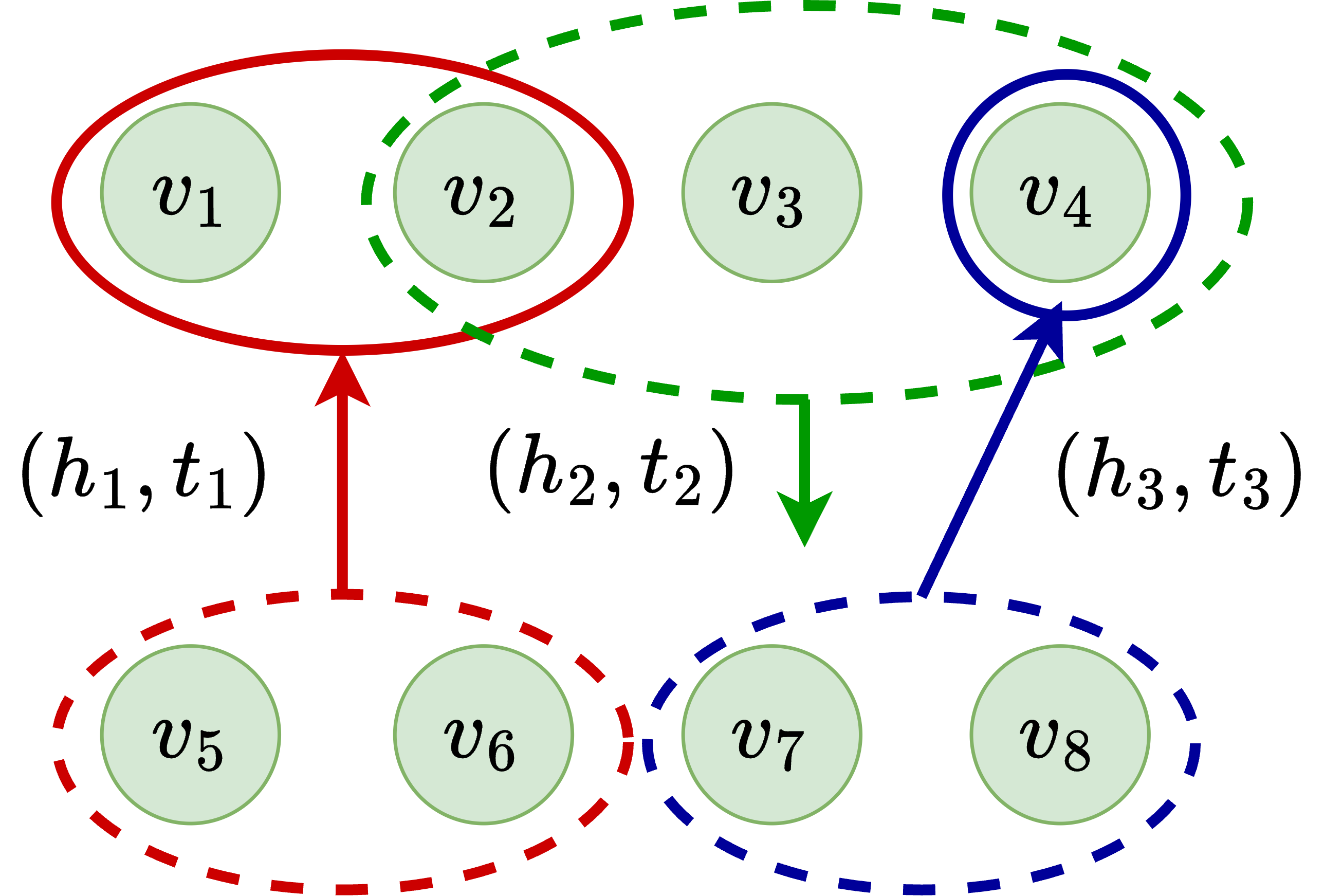}
    \caption{Model}
    \label{fig:txn_model}
  \end{subfigure}
  \caption{Bitcoin transactions (Txns) are modeled as a temporal directed hypergraph graph with eight nodes to represent addresses, and three hyperedges to represent transactions. Here, $t_i$ is the time with $t_3 > t_2 > t_1$, and a hyperedge is represented as a tuple $ h_i = (h_i^r, h_i^\ell)$ with $h_i^r$ the right hyperedge are the senders' addresses and $h_i^\ell$ the left hyperedge are receiver addresses. }
  \label{fig:bitcoin_network_model}
\end{figure*}

An example of relationships in a Bitcoin transaction network as directed hyperedges is depicted in Figure~\ref{fig:bitcoin_network_model}. Here, we can see that Bitcoin transactions involve two groups of entities: one group representing the senders and another group representing the receivers. The goal is to create models that can track the evolution of users in the Bitcoin network, represented by addresses, over time so that we can forecast transactions. Using learned dynamic representations, these models can then be used to detect anomalous transactions occurring within the network, such as money laundering~\citep{WuEtAL:2022:DetectingMixingServicesViaMiningBitcoinTransactionNetworkWithHybridMotifs}. 


Existing works on directed hyperedge prediction use hypergraph neural networks based unsupervised learning~\citep{WeiEtAL:2022:AugmentationsInHypergraphContrastiveLearningFabricatedAndGenerative,LeeEtAL:2023:ImMeWeReUsAndImUsTridirectionalConstrastiveLearningOnHypergraphs}. These models cannot be applied to the proposed problem as hypergraphs cannot be constructed from historical relations as they exist only for an instant of time. Alternately, set prediction-based scoring functions are used for hyperedge prediction. These models use a deep learning-based permutation invariant architecture to make predictions from node representations. These works are applicable only for undirected hyperedges~\citep{ZhangEtAL:2019:Hyper-SAGNNASelfAttentionBasedGraphNeuralNetworkForHypergraphs}.
These models can capture only self-connection information in the case of the undirected and cross-connection in the case of the bipartite. Unlike these, a directed hyperedge can represent three types of information, two self-connections among the left and right groups and a cross-connection between these groups. Now to model the evolution of nodes, we use temporal node representations. Previous work on higher-order relation modeling uses it for forecasting undirected and bipartite hyperedges events~\citep{GraciousEtAL:2023:DynamicRepresentationLearningWithTemporalPointProcessesForHigherOrderInteractionForecasting}. These works achieve this using a sequential recurrent neural network-based model that updates node representations when an event occurs. One of the main disadvantages of this approach is that it will not allow the model to do batch processing of the data, as each sample depends on the previous samples. Further, for simulating the future in directed temporal graphs with nodes $\calV$, one needs to compare all the ${}^{|\calV|}\calP_{2}$ pairwise combinations of nodes in the network. In the case of directed hypergraphs, there are a maximum of $2^{|\calV|}$ combinations of nodes in the left and the right hyperedges.

In this paper, for the first time, we solve the problem of learning the representation of a higher-order, directional, and temporal network by proposing a model called \textit{Directed HyperNode Temporal Point Process}, \textbf{DHyperNodeTPP}. Unlike previous works that forecast future events by an exhaustive search, we use a generative model based on the Temporal Point Process (TPP) to forecast candidate hyperedges. This is achieved by predicting event times for nodes followed by forecasting the projected adjacency matrix along with the distribution of hyperedge sizes of the nodes that are then used for generating candidate hyperedges. Further, to process batches of data, we use a temporal message-passing technique to store the features of events on each node and then use them to update the node representation before the next batch. Appendix \ref{sec:appendix:scalable_training} shows the advantage of using batch processing to reduce the training time, enhancing model scalability to datasets with many samples. The following are the main contributions of our work. 
\textbf{(1)} A temporal point process model for forecasting directed hyperedges in a scalable way;  
\textbf{(2)} A temporal node representation learning approach that uses graph attention networks and batch processing; 
\textbf{(3)} A directed hyperedge prediction model; 
\textbf{(4)} Creation of five real-world temporal-directed hypergraph datasets from open-source data; 
\textbf{5)} Extensive experiments showing the advantage of our model on event forecasting over existing modeling techniques. 

\paragraph{Related Works.}
The early works in temporal relation forecasting focus only on pairwise edge prediction. These involve discrete time models like DySAT~\citep{SankarEtAL:2020:DySATDeepNeuralRepresentationLearningOnDynamicGraphsViaSelfAttentionNetworks} and NLSM~\citep{GraciousEtAL:2021:NeuralLatentSpaceModelForDynamicNetworksAndTemporalKnowledgeGraphs}, which divide the time into snapshots of uniform length and predict edges in the future snapshot by observing the history. Our method focuses on building continuous time models since discretization involves information loss and requires domain knowledge. JODIE~\citep{KumarEtAL:2019:PredictingDynamicEmbeddingTrajectoryInTemporalInteractionNetworks} and TGAT~\citep{XuEtAL:2020:InductiveRepresentationLearningOnTemporalGraphs} use temporal graph network to aggregate history into node representations, and these are trained to predict edge formation as a binary classification. Even though they can predict the presence of an edge at a particular time, they cannot estimate the time at which the event will occur. The TPP is a popular technique for modeling temporal networks that can do both. DyRep~\citep{TrivediEtAL:2019:DyRepLearningRepresentationsoverDynamicGraphs} and DSPP~\citep{CaoEtAL:2021:DeepStructuralPointProcessForLearningTemporalInteractionNetworks} use the TPP to model the time distribution of edge formation. Note that none of these above-mentioned works can model higher-order relations available abundantly in the real world. The importance of higher-order analysis is empirically shown in the work by~\citet{BensonEtAL:2018:SimplicialClosureAndHigherOrderLinkPrediction}. They focus on the problem of higher-order relations between three nodes, mainly on the issue of distinguishing between the formation of the open triangle and close triangle relations.  This is extended to a deep learning based method by \citet{LiuEtAL:2021:NeuralHigherOrderPatternMotifPreidctionInTemporalNetworks}. 
Recently, \citep{GraciousEtAL:2023:DynamicRepresentationLearningWithTemporalPointProcessesForHigherOrderInteractionForecasting} have addressed the problem of hyperedge forecasting using a TPP model, but this work does not consider the scalability issues and direction information in the relations.

\section{Problem Definition}
\label{sec:problem_definition}
\begin{figure}
    \centering
    \includegraphics[width=\linewidth]{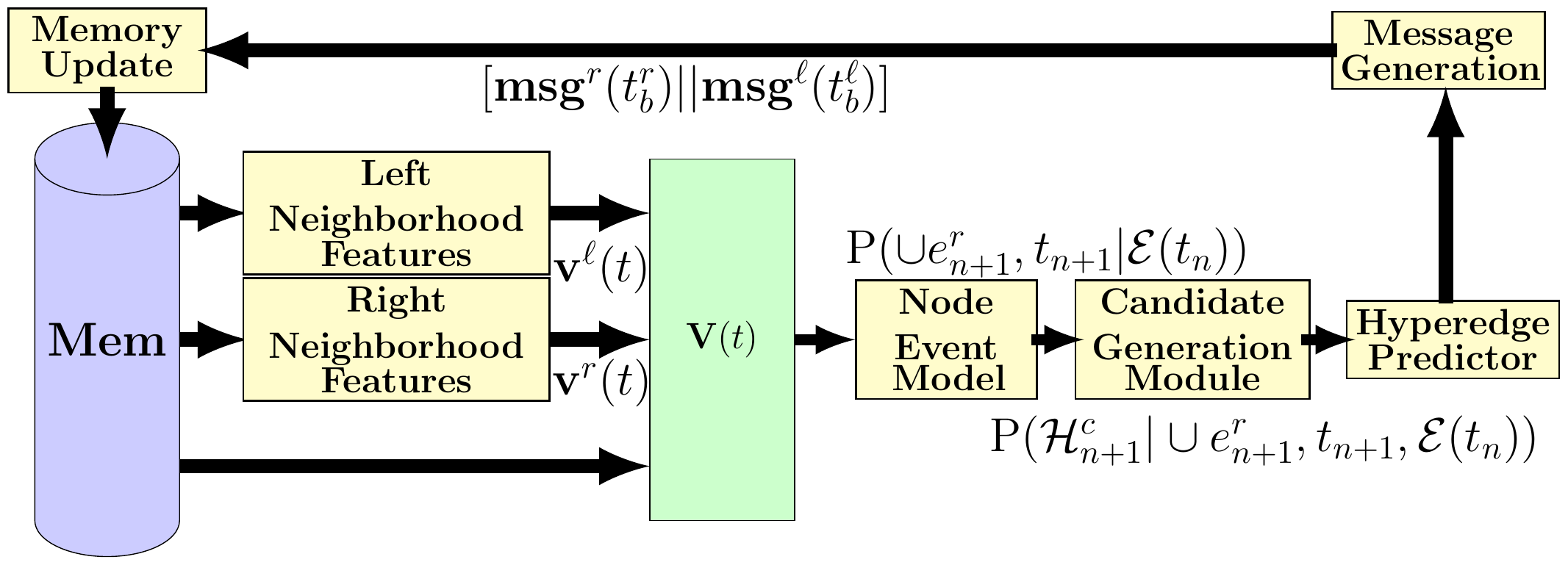}
    \caption{Neural Architecture of DHyperNodeTPP: We calculate the temporal node representation $\bfV(t)$ by combining the entries from the Memory module with information from recent relations where the node is involved in the left and right hyperedges. These temporal node representations are given as input to forecast nodes where events occur, followed by candidate hyperedge generation. Then the hyperedge prediction decoder  in Section \ref{sec:architecture_of_link_predictor} is used to find the observed hyperedges.}
    \label{fig:model}
\end{figure}

\paragraph{Hypergraph.} A directed hypergraph is denoted by $\calG = ( \calV, \calH )$, where $\calV = \{v_1, v_2, \ldots, v_{|\calV|}\}$ is the set of nodes, $\calH$ is the set of valid hyperedges. In this, each hyperedge $h=(h^r, h^\ell) \in \mathcal{H}$  is represented by two subset of nodes, $h^r \subset \calV$ the right hyperedge of size $k^r= |h^r|$, and $h^\ell \subset \calV$ the left hyperedge of size $k^\ell=|h^\ell|$. The maximum size of right and left hyperedge is denoted by $k^r_{max} = \max \limits_{h \in \calH} | h^r|$ and $k^\ell_{max} = \max \limits_{h \in \calH} | h^\ell|$, respectively. 
\paragraph{Temporal Events.} The sequence of events occurring in a hypergraph till time $t \in \bbR^{+} $ is denoted by  $ \calE (t)=\{(e_{1}, t_{1}), \ldots, (e_n, t_n), \ldots \}$. Here, event $e_n= \{ h_{n,m}\}_{m=1}^{L_n}, h_{n,m} \in \calH $, denotes the $L_n$ concurrent hyperedges occurring at time $t_n \leq t$  with $n$ as the index of the event and $m$ as the index of the concurrent hyperedge. Then for each $e_n$, we create its projected adjacency matrices $ \bfA^r_n, \bfA^\ell_{n}  \in {\{0,1\}}^{|\calV| \times |\calV| }$  indicating the pairwise adjacency matrices for nodes in right hyperedge. In these, $\bfA^r_n[i,j]=1$ if $\exists\ \{v_i, v_j\} \subset h^r,  h=(h^r, h^\ell) \in e_{n}$ else $0$, and   $\bfA^\ell_{n}[i,j]=1$ if $\exists\ v_i \in h^r$ and $\ v_j \in h^\ell, h=(h^r, h^\ell) \in e_{n}$ else $0$. The rows of $ \bfA^r_n$ and $\bfA^\ell_{n}$  denoted by $\bfa^r_{n,i}$ and $\bfa^\ell_{n, i}$, respectively, are the adjacency vectors of node $v_i$. In addition to this, we create the size matrices of right $\bfK^r_n  \in {\{0,1\}}^{\calV \times k^\ell_{max}  } $  and left hyperedges $\bfK^\ell_n  \in {\{0,1\}}^{|\calV| \times k^r_{max} } $ with respect to the nodes in right hyperedge. Here, $ \bfK^r_{n} [i, k]=1$ if $\exists  k=|h^r|$ and  $\bfK^\ell_{n} [i, k]=1$ if $\exists  k=|h^\ell|$  $\forall h \in e_{n+1}$ and $v_i \in h^r$. The size vectors for a node $v_i$ is denoted by $\bfk^\ell_{i,n}$ and $\bfk^r_{i,n}$. These are the $i$th rows of respective matrices. 
%
%
%
%
%
\paragraph{Goal.} The aim is to learn the  probability distribution $\rmP^{*}( e_{n+1}, t_{n+1} )  = \rmP( (e_{n+1}, t_{n+1})  | \calE (t_n))$ over the time ($t_{n+1}$), and type ($e_{n+1}$) of  the next event given the history, $ \calE (t_n)$. 
A naive implementation has the following likelihood, 
$
\rmP^{*}( e_{n+1}, t_{n+1} ) =  \rmP^{*}(t_{n+1} ) \prod_{h \in \calH} [ \rmP^{*}( h |  t_{n+1}) ]^{\bbI_{h \in e_{n+1}}} \times  
[ 1 - \rmP^{*}( h |  t_{n+1} ) ]^{\bbI_{h \notin e_{n+1}}}.
$
This is obtained by assuming all the relations inside event $e_{n+1}$ are independent, given the time of the event and history. 
The main challenge in generating samples based on the above likelihood is that the hyperedge prediction requires a search over a huge number of candidates, $|\calH|$. Alternately, we can further reduce this by introducing a candidate generation model that will forecast likely hyperedges ($\calH^c_{n+1}$), 
\begin{align}\label{eq:complete_gen_process}
    &\rmP^{*}( e_{n+1},  t_{n+1} )  \nonumber \\
    &=\rmP^{*}(  \cup e_{n+1}^r, t_{n+1}  ) \times\rmP^{*}( \calH^c_{n+1} | \cup e_{n+1}^r, t_{n+1}   ) \times \nonumber \\  
    &\prod_{h \in \calH^c_{n+1}} [ \rmP^{*}( h |  t_{n+1}  ) ]^{\bbI_{h \in e_{n+1}}} [ 1 - \rmP^{*}( h |  t_{n+1}  ) ]^{\bbI_{h \notin e_{n+1}}} .
\end{align}
Let $\cup e_{n+1}^r = \cup_{m=1}^{L_{n} }  h_{n+1,m}^r $ be the set of all the nodes in the right hyperedge. In this model, we first predict all the right nodes that are observing events at time $t_{n+1}$, followed by a candidate hyperedge generation module that outputs $\calH^c_{n+1}$, and finally, we use the directed hyperedge predictor to get the ground truth. Here, we assume that the right hyperedge represents the source of the relation; for email exchange, the right node is the sender, and citation networks have the right nodes as the paper's authors. The entire block diagram for the proposed model is shown in Figure \ref{fig:model}.
%
%

\section{Model}
\label{sec:model}
We follow a sequential generative process by dividing the relation forecasting into three different modules that do time forecasting, candidate generation, and hyperedge prediction. These models use the temporal representations of nodes $\bfV (t)  \in \bbR^{|\calV| \times d}$ as input. Here $d$ is the dimension of representation, and $\bfv_i(t) \in \bbR^d$ denotes the representation of node $v_i$.  The architecture used to learn this is explained in Section \ref{sec:dynamic_node_embedding}. In the following section, $\MLP_{*}$s are multilayer perceptron functions, and architectures for these are explained in Appendix \ref{sec:mlp_layers_appendix}.
\paragraph{Node Event Model.}
%
Given the history, $\mathcal{E}(t_n)$, the task is to model the probability distribution of the time of occurrence of events on nodes. This is the product of likelihood of next event occurring at time $t_{n+1} = \Delta t_{n+1} + t_{n}$  for nodes in $\cup e_{n+1}^r$ and event not occurring for nodes not in $\cup e_{n+1}^r$ for the interval $[t_n, t_{n+1}]$.
$
     \rmP^{*}( \cup e_{n+1}^r, t_{n+1} ) =  \prod_{v_i \in \cup e_{n+1}^r} \rmP^{*}_i ( \Delta t_{n+1}   ) \prod_{v_i \notin  \cup e_{n+1}^r} \rmS^{*}_i ( \Delta t_{n+1}   ).
$
Here, $\rmS_i^{*}$ is the survival function that models the probability of events not occurring in the interval $[t_{n}, t_{n+1}]$.
We use $\mathrm{Lognormal}$ distribution to model event times, $\Delta t_{n+1} \sim \mathrm{Lognormal}(\mu_{n+1, i }, s_t^2)    $.  Here, variance $s_t$ is a hyperparameter, and $\mu_{n+1, i}$ is parameterized by a neural network that takes representation of $v_i$ at time $t_n$ as the input, $\mu_{n+1, i} = \MLP_t (\bfv_i(t_n) )$. Then the log-likelihood becomes,
\begin{align*} 
    \calL \calL_{t}^{n+1} &= \sum_{v_i \in \cup e_{n+1}^r}   \frac{( \log(\Delta t_{n+1})  - \mu_{n+1, i}) ^2}{2 s_t^2}  \nonumber \\ &
    - \sum_{v_i \notin \cup e_{n+1}^r}  \log \left( 1 - \Phi \left( \frac{ \log\Delta t_{n+1}  - \mu_{n+1, i} }{s_t} \right) \right).
\end{align*}
Here, $\Phi(.)$ is the cumulative density function of the standard normal. The second loss component due to the survival function is approximated using negative sampling during training.

\paragraph{Candidate Generation.} This is achieved by predicting the projected adjacency,  $ \bfa_{n+1,i}^r, \bfa_{n+1,i}^\ell$, and size vectors, $ \bfk_{n+1,i}^r, \bfk_{n+1,i}^\ell$,  for node $v_i$ observing event from previous module, $v_i \in \cup e_{n+1}^r$. We use independent $\mathrm{Bernoulli}$ distribution to model the adjacency vectors of $v_i$, 
$
 \bfa^r_{n+1, i}  \sim \mathrm{Bernoulli}(\sigma( \bmtheta_{n+1,i}^r ) ), \bfa^\ell_{n+1, i} \sim \mathrm{Bernoulli}( \sigma( \bmtheta_{n+1,i}^\ell )  ).
$
Here, $\bmtheta^r_{n+1,i}  = \MLP_{ar} (\bfv_i(t_n) )$ and $\bmtheta^\ell_{n+1,i} = \MLP_{a \ell} (\bfv_i(t_n) )$. Similarly, we model the size distribution for node $v_i$,
$
    \bfk^r_{n+1,i} \sim \mathrm{Bernoulli}(\sigma( \bmkappa^r_{n+1,i}))  ,  \bfk^\ell_{n+1,i}  \sim \mathrm{Bernoulli}( \sigma(\bmkappa^\ell_{n+1,i}  ) ) .
$
Here, $ \bmkappa^r_{n+1,i} = \MLP_{sr} (\bfv_i(t_n)) $, and  $\bmkappa^\ell_{n+1,i} =  \MLP_{sl} (\bfv_i(t_n) )  $.  Then we can write log-likelihood for size and adjacency prediction as,
\begin{align}
    \label{eq:size_loss}
    \calL \calL_{k}^{n+1} &=   \sum_{v_i  \in \cup e_{n+1}^r, s={r,l} } \langle \bfk^{s}_{n+1,i}  , \log( \sigma(\bmkappa^s_{n+1,i} ) ) \rangle  \nonumber \\ &+   \langle (\bm{1} - \bfk^{s}_{n+1,i} ), \log(\bm{1} - \sigma( \bmkappa^s_{n+1,i} ) ) \rangle,
\end{align} 
\begin{align}
    \label{eq:adjacency_loss}
    \calL \calL_{a}^{n+1} &=   \sum_{v_i  \in \cup e_{n+1}^r , s={r,\ell}} \langle \bfa^s_{n+1, i}, \log( \sigma( \bmtheta^s_{n+1,i}  ) \rangle  \nonumber \\ & +  \langle (\bm{1} -  \bfa^s_{n+1, i} ) , \log(\bm{1} - \sigma( \bmtheta^s_{n+1,i}  ) ) \rangle .
\end{align} 

%
\paragraph{Hyperedge Predictor.} Given the candidate hyperedges $\calH^{c}_{n+1}$, the the probability of observing $e_{n+1}$ at time $t_{n+1}$ is,
$\rmP^*(  e_{n+1} | \calH^{c}_{n+1}, t_{n+1}) = \prod_{ h \in \calH^{c}_{n+1} } \sigma ( \lambda_h  (t_{n+1}) )^{\bbI_{h \in e_{n+1}}}    (1-\sigma( \lambda_{h}(t_{n+1})) )^{\bbI_{h \notin e_{n+1}}} $. 
Here, $\lambda_{ h}(t)$ is parameterized by a neural network that takes representations of the nodes in $h$ at time $t$ as input,
$
 \lambda_{h} (t) = f( \{ \bfv_{i} (t)\}_{v_{i} \in  h^r }, \{ \bfv_{i} (t)\}_{v_{i} \in  h^\ell} ).
$
The architecture of $f(.)$ is explained in Section \ref{sec:architecture_of_link_predictor}. The log-likelihood is,
\begin{align}
    \calL \calL_{h}^{n+1}  = & \sum_{ h \in \calH^{c}_{n+1} } {\bbI_{h \in e_{n+1}}} \log \sigma ( \lambda_h  (t_{n+1}) ) \nonumber \\ 
  &   +    {\bbI_{h \notin e_{n+1}}} \log (1-\sigma( \lambda_{h}(t_{n+1})) ).
\end{align}
While training, the candidate hyperedges are the combination of the true hyperedges and negative hyperedges generated by negative sampling. This is done by replacing either the left or right of true hyperedge with a hyperedge of randomly sampled size and filled with random nodes.
\paragraph{Loss Function.} The complete likelihood for event sequence $\calE (t_N)$ is, $    \rmP(  \calE(t_N) ) = \prod_{n=0}^{N-1} \rmP^{*}( e_{n+1}, t_{n+1}  )$.  For training, we minimize the negative log-likelihood, 
$$
\calN \calL \calL  = -\left[ \sum_{n=0}^{N-1} \calL \calL_{t}^{n+1} + \calL \calL_{k}^{n+1} + \calL \calL_{a}^{n+1} + \calL \calL_{h}^{n+1} \right].
$$


\subsection{Architecture of Hyperedge Predictor}\label{sec:architecture_of_link_predictor}
We now propose an architecture that utilizes all three types of information in a directed hyperedge to predict true hyperedges from candidate hyperedges.  Given a directed hyperedge $h = (h^r, h^\ell)$ with $h^r=\{v_{1^{r}}, \ldots, v_{k^{r}} \}$ and $h^\ell = \{ v_{1^{\ell}}, \ldots, v_{k^{\ell}}  \} $, we use the cross-attention layer ($\mathrm{CAT}$) between the sets of nodes to create cross-dynamic hyperedge representation, $$\bfd^{ch} _{i^r}  = \mathrm{CAT} ( \{ \bfv_{1^{r}}(t), \ldots, \bfv_{k^{r}} (t)  \}, \{ \bfv_{1^{\ell}}(t), \ldots, \bfv_{k^{\ell}}(t)  \}  ),$$ for node $v_{i^r}$. These dynamic hyperedge representations are added to the original representation to get $\bfz_{i^r}(t) = \bfv_{i^r}(t) + \bfd^{ch }_{i^r}$. Then we pass these through a self-attention layer ($\mathrm{SAT}$) to get self-dynamic hyperedge representation, $\bfd^{sh}_{i^r}  = \mathrm{SAT} ( \{ \bfz_{1^{r}}(t), \ldots, \bfz_{k^r}(t)  \}  )$. The complete dynamic hyperedge representations are obtained by combining them, $\bfd^{h}_{i^r}=\bfd^{sh}_{i^r} + \bfd^{ch}_{i^r} $. 
We also create static hyperedge representation $\bfs^h_{i^r} = \bfW_{s^r} \bfv_{i^r}(t)$, where $\bfW_{s^r} \in \mathbb{R}^{d \times d}$ is a learnable parameter. Note that the terms dynamic and static in this section are used with respect to the hyperedge, not the time. Then we calculate the Hadamard power of the difference between static and dynamic hyperedge representation pairs followed by a linear layer, and the final score $\calP^{h^r}$ is calculated as shown below,
\begin{align*}
    & \calP^{h^r}   =  \frac{1}{k^r}\sum_{{i^r}=1}^{k^r}   \bfW_{o^r}  [ (\bfd^h_{i^r} - \bfs^h_{i^r}   ) ^{2} || (\bfd^{ch}_{i^r} - \bfs^h_{i^r}   ) ^ 2 ]  + b_{o^r}.
\end{align*}
Here, $||$ is the concatenation operator, $\bfW_{o^r} \in \mathbb{R}^{ 1 \times 2d}, b_{o^r} \in \mathbb{R}$ are the learnable parameters of the output layer. 
This equation models the cross and self connections in the right hyperedge. Similarly, we model left hyperedge to find $\mathcal{P}^{h^\ell}$  with a different set of parameters and combine them to predict links, $f( \{ \bfv_{i} (t)\}_{v_{i} \in  h^r }, \{ \bfv_{i} (t)\}_{v_{i} \in  h^\ell } )= \mathcal{P}^{h^r} + \mathcal{P}^{h^\ell}$. 
The model's entire block architecture and details of $\mathrm{CAT}$ and $\mathrm{SAT}$ are shown in Appendix \ref{sec:attention_layers_appendix}. 

\subsection{Temporal Node Representation} 
\label{sec:dynamic_node_embedding}
The temporal node representation of node $v_i$ has the following form,
$
    \bfv_i(t) =  \mathrm{tanh}(\bfW_s \mathbf{Mem}_{i}  + \bfW^r  \bfv^r_i(t)  +  \bfW^\ell \bfv^\ell_i (t) + \bfb_v ).
$
Here, $ \bfW_s \in \mathbb{R}^{d \times d }, \bfW^r \in \bbR^{d \times d },   \bfW^\ell \in \mathbb{R}^{d \times d }$ and $\bfb_v \in \mathbb{R}^d$ are learnable parameters. The first term, $\mathbf{Mem}_{i}\in \mathbb{R}^d$, is the historical events information for node $v_i$ stored in the Memory Module, as explained in the following section. The second term, $\bfv^r_i(t)$, is the right neighborhood features calculated based on the recent higher-order relations $h$ where the node $v_i$ is present in the right hyperedge, $v_i \in h^r$. Similarly,  we calculate $\bfv^l_i(t)$ based on the relations where a node is present in the left hyperedge. The architectures used for each of these components are shown below.
\paragraph{Memory Module.}%
$\mathbf{Mem}  \in \mathbb{R}^{|\mathcal{V}| \times d }$ stores the historical event information for each node in the network till time $t$. This module is initialized with zero values, and when a node is involved in a relation, its corresponding entries are updated based on the output of the Message Generation Module. However, if we update the memory with each event as done in the previous work, \citep{GraciousEtAL:2023:DynamicRepresentationLearningWithTemporalPointProcessesForHigherOrderInteractionForecasting}, we will not be able to scale the model to a large number of samples, as batch processing of the data is not possible. So, we divide the events into batches of size $\calB$ while maintaining their temporal order. Then, we aggregate the relation features of each node in the batch using the message generation stage, which is used to update the memory entries of the node using a recurrent neural network, as explained in the memory update stage.
\paragraph{Message Generation.} 
We calculate features for each node in all the hyperedges in a batch ${(h_b, t_b)}_{b=1}^{\calB}$. For node $v_i$ in the right hyperedge $h^r_b$ of $(h_b, t_b)$, the features are the concatenation of its temporal node representation $\bfv_i(t)$, dynamic hyperedge representation $\bfd^h_{i}$, and Fourier features \citep{XuEtAL:2020:InductiveRepresentationLearningOnTemporalGraphs}  calculated from the duration $ t- t_{i}^p $ since the last memory update of that node at time $t_{i}^p $. This is stored as a message vector, 
$
    \mathbf{msg}^r_{i}(t_b ) = \left[\bfv_i(t) || \bfd^h_{i} || \bmpsi (t- t_{i}^p) \right]  .
$
 Here, $\bmpsi ( t - t_{i}^{p}) \in \mathbb{R}^{d}$ is the Fourier features based on functional encoding for the time $t - t_{i}^{p}$. This is achieved by learning a mapping function,
$
        \bmpsi (t) = [\cos(\omega_1 t + \phi_1) , \ldots, \mathrm{cos}(\omega_d t + \phi_d)  ]    
$
with parameters $\{\omega_i \}_{i=1}^d$, and  $\{\phi_i \}_{i=1}^d$ inferred from the data. Similarly, we calculate the message vectors $(\mathbf{msg}^{l}(.))$  for nodes in the left hyperedge. 
\paragraph{Memory Update.} 
The messages from the previous section are used to update the memory entries of the corresponding nodes before the next batch. In our work, we use a $\mathrm{GRU}$ \citep{ChoEtAL:2014:LearningPhraseRepresentationsUsingRNNEncoderDecoderForStatisticalMachineTranslation} based recurrent neural networks for memory updating as,
$
    \mathbf{Mem}_{i} = \mathrm{GRU}([ \mathbf{msg}^{r}_{i}(t^{r}_{b} )|| \mathbf{msg}^{\ell}_{i}(t^{\ell}_{b} ) ],  \mathbf{Mem}_{i} ),
$ and  $t_{i}^p =  max( t^r_b, t^\ell_b)$. Here, $t^{r}_{b},t^{\ell}_{b} $ are the latest event times for node $v_i$ in previous batch. The corresponding features will have zero value for a node if the messages are unavailable on the left or right.
%
\paragraph{Neighborhood Features.} 
To incorporate higher-order neighborhood information into the node representation, we use a graph attention network to update the node representation with features from relevant historical events.  Further, it also helps to avoid the staleness of vectors in the Memory Module due to the absence of recent events involving the node by extracting information from other nodes that are previously involved with it \citep{KazemiEtAL:2020:RepresentationLearningForDynamicGraphASurvey}.
For a node $v_i$ at time $t$, we find the recent $\calN$ relations involving the node in the right hyperedge, and we denote them as $\calN_{h^r}(t)$.  Then for each hyperedge $(h, t) \in \calN_{h^r}(t)$, we calculate the hyperedge representation as shown below,
$
    \bfh^r(t_i) = \frac{1}{|h^r|} \sum_{v_{i^r} \in h^r}\bfW^r_h \mathbf{Mem}_{i^r} +  \frac{1}{|h^\ell|}  \sum_{v_{i^\ell} \in h^\ell} \bfW^\ell_h \mathbf{Mem}_{i^\ell}  .
$
Similarly, we find the recent $\calN$ relations where nodes are in the left hyperedge, $\calN_{h^\ell}(t)$, and calculate hyperedge representation $\bfh^\ell(t_i)$. Then, the right neighborhood features are calculated using the graph attention layer as shown below,
\begin{align}
     &\bfC(t)  = \left[ \bfh^r(t_1) ||\bmpsi(t-t_1), \ldots,   \bfh^r(t_{\calN}) ||\bmpsi(t-t_{\calN}) \right], \nonumber\\ 
    &\bfq (t)  = \mathbf{Mem}_{i} || \bmpsi(0),   \nonumber \\ 
    &\bfv^r(t) = \mathrm{MultiHeadAttention}(\bfq (t), \bfC (t), \bfC (t)) .
\end{align}
 The $\mathrm{MultiHeadAttention}$ uses the node memory vectors for query ($\bm{\rmq}(t)$), and recent relation representation as keys ($\bfC (t) $) and values ($\bfC (t)$)~\citep{VaswaniEtAL:2017:AttentionIsALLYouNeed}. Similarly, we use a separate $\mathrm{MultiHeadAttention}$ layer to calculate $\bfv^\ell(t)$.

\section{Experiments}
\label{sec:experiments}
%
\begin{table}
\centering
\small 
\setlength{\tabcolsep}{2pt}
\begin{tabular}{lccccc}
\toprule
\textbf{Datasets}  &    $|\mathcal{V}|$  & $|\mathcal{E}(T)|$ & $|\mathcal{H}^r|$ & $|\mathcal{H}^{\ell}|$  & $T$  \\
\midrule 
\textbf{Enron-Email} & 183  & 10,311 & 1,003 & 89 & 99,070\\
\textbf{Eu-Email} &  800  & 208,403 &  11,897 & 744 & 69,459,254  \\
\midrule
\textbf{Twitter} &  2,130 & 9,889 & 1,218 & 2,321 & 17,277 \\
\midrule 
\textbf{Hepth}  & 451 & 9,882 & 8,384 & 1,352 & 21,532 \\
\textbf{ML-Arxiv}  & 659 & 18,558  & 2,995 & 17,014 & 62,741 \\
\midrule
\textbf{Bitcoin} &  7,806 & 231,071 & 23,901 & 6,706 & 1,416,317,420  \\ 
 \midrule 
\textbf{iAF1260b} & 1,668 & 2,084  & 2,010 & 1,985  & $N/A$ \\
\textbf{iJO1366} & 1,805 & 2,253 & 2,174 & 2,146  & $N/A$\\
\textbf{USPTO} & 16,293 & 11,433 & 6,819 & 6,784 & $N/A$ \\
\bottomrule
\end{tabular}
\caption{Datasets used for Temporal and Static Directed Hypergraphs along with their vital statistics.}
\label{tab:datasets}
\end{table}

\paragraph{Datasets.} 
%
%
Table \ref{tab:datasets} shows the statistics of both static and temporal directed hypergraph datasets. Here, $|\mathcal{V}|$ denotes the number of nodes, $|\mathcal{E}(T)|$ denotes the number of hyperedges, $|\mathcal{H}^{r}|$ denotes the number of unique right hyperedges, $|\mathcal{H}^{l}|$ denotes the number of unique left hyperedges, and $T$ is the time span of the dataset. For the static datasets, there is no time span feature.  A more detailed description of each dataset is provided in Appendix \ref{sec:datasets_description_appendix}.
\paragraph{Baseline Models.}
HyperNodeTPP is an undirected version of our model DHyperNodeTPP with right and left hyperedge merged into a single hyperedge, $h = h^r \cup h^\ell $. This uses the same temporal node representations with only a single type of neighborhood features and  HyperSAGNN architecture for hyperedge prediction~ \citep{ZhangEtAL:2019:Hyper-SAGNNASelfAttentionBasedGraphNeuralNetworkForHypergraphs}. 
HGBDHE and HGDHE are models taken from previous work for higher-order relation forecasting by~\cite{GraciousEtAL:2023:DynamicRepresentationLearningWithTemporalPointProcessesForHigherOrderInteractionForecasting}.  TGN and GAT are pairwise models developed for forecasting pairwise relations. TGN is developed for forecasting edges in a temporal graph~\citep{RossiEtAL:2020:TemporalGraphNetworksForDeepLearningOnDynamicGraphs}, and GAT~\citep{VeličkovićEtAL:2018:GraphAttentionNetworks} forecast edges in a static graph.
%
%
%
%

\begin{table*}[!htbp]
    \centering
    \small 
     \setlength{\tabcolsep}{3pt}
    \begin{tabular}{llccccccc}
        \toprule
        \multicolumn{2}{}{}{\textbf{Methods}}  & \textbf{GAT}&  \textbf{TGN} & \textbf{HGDHE} & \textbf{HGBDHE} & \textbf{HyperNodeTPP}(ours) & \textbf{DHyperNodeTPP}(ours) \\ 
        \midrule
        \multirow{2}{*}{\textbf{Enron-Email}} & MAE & $N/A$ & $N/A$  & 35.77 $\pm$ 1.61 & 13.92 $\pm$ 0.35 &  $\bm{ 4.15 \pm 0.01 }$ & 4.18 $\pm$ 0.02  \\
        & MRR  & 40.16 $\pm$ 7.15 & 42.22 $\pm$ 0.87 & 35.00 $\pm$ 3.94 & 36.09 $\pm$ 1.96 & 61.85 $\pm$ 0.01 & $\bm{61.94 \pm 0.01}$ \\
        \multirow{2}{*}{\textbf{Eu-Email}} & MAE & $N/A$ & $N/A$  &  20.58 $\pm$ 3.34  & 18.57 $\pm$ 1.62 &  12.23 $\pm$ 0.03 &  $\bm{12.22 \pm 0.02}$ \\   
           & MRR & 66.81 $\pm$ 0.02 & 69.15 $\pm$ 0.01 & 62.42 $\pm$ 1.79 & 55.34 $\pm$ 1.21 &  $\bm{75.95 \pm 0.01} $  & 68.05 $\pm$ 0.03  \\
        \midrule
        \multirow{2}{*}{\textbf{Twitter}} & MAE & $N/A$ & $N/A$ & 21.58 $\pm$ 3.79 & 8.16 $\pm$ 0.70 & $\bm{1.18 \pm 0.01}$ & 1.20 $\pm$ 0.01 \\
        & MRR  & 44.88 $\pm$ 0.05  &  55.20 $\pm$ 0.03 & 69.87 $\pm$ 0.72 & 70.19 $\pm$ 0.95 &  $\bm{84.47 \pm 0.01}$ & 82.12 $\pm$ 0.00 \\
        \midrule
        \multirow{2}{*}{\textbf{HepTh}} & MAE & $N/A$ & $N/A$ & 16.19 $\pm$ 3.01 & 8.86 $\pm$ 0.08 & 1.25 $\pm$ 0.03 & $\bm{1.20 \pm 0.01}$ \\
        & MRR & 33.79 $\pm$ 8.95  & 51.70 $\pm$ 1.37 & 57.98 $\pm$ 0.82 & 57.40 $\pm$ 3.00 & 45.18 $\pm$ 0.02 & $\bm{79.01 \pm 0.01}$ \\
        \multirow{2}{*}{\textbf{ML-Arxiv}} & MAE &  $N/A$ & $N/A$  & 29.94 $\pm$ 3.77 & 17.29 $\pm$ 0.57 & 1.25 $\pm$ 0.00 & $\bm{1.24 \pm 0.00}$ \\
        & MRR & 22.49 $\pm$ 4.31  & 37.58 $\pm$ 1.12 & 26.07 $\pm$ 0.29 & 28.13 $\pm$ 0.78 & 29.49 $\pm$ 0.00 & $\bm{52.05 \pm 0.01}$ \\
        \midrule 
        \multirow{2}{*}{\textbf{Bitcoin}} & MAE &  $N/A$ & $N/A$ & 109.71 $\pm$ 4.34 &  $\bm{63.16 \pm 1.55}$ &   75.98 $\pm$
1.44 & 75.73 $\pm$
2.60   \\
            & MRR  & 93.61 $\pm$ 4.02 &  93.34 $\pm$ 1.48 & 91.03 $\pm$ 0.00  & 85.06 $\pm$ 0.03 & 97.50  $\pm$ 0.09  &  $\bm{98.72 \pm
0.05}$\\
        \bottomrule
    \end{tabular} 
    \caption{Results in event type and time prediction tasks. The proposed model DHyperNodeTPP beats baseline models in almost all the tasks. Here, event type prediction is evaluated using MRR $\%$; here, a higher value indicates better performance, and event time prediction is evaluated using MAE, the lower value indicating better performance.}
    \label{tab:dynamicdirectedlinkprediction}
\end{table*}

%

%

We create two tasks to evaluate our model's event forecasting capabilities and to compare its performance against baseline methods. (i) Event Type Prediction: The goal of this task is to predict the type of event $e_{n+1}$ occurring at time $t_{n+1}$ given the history $\mathcal{E}(t_n)$.  For evaluating the performance, we find the position of true hyperedges against candidate negative hyperedges by ordering them in descending order of $\lambda_h(t)$. Here, negative hyperedges are calculated by replacing the entire left or right hyperedges with hyperedges of randomly sampled nodes and sizes.  Then MRR is calculated as follows, 
$
    \text{MRR} = \frac{1}{N} \sum_{n=1}^{N} \frac{1}{r_n + 1}.
$
(ii) Event Time Prediction: The goal of this task is to predict the next time of event $t_{n+1}$  given the history $\mathcal{E}(t_n)$ for the nodes of interest. It can be calculated using the time estimate $\Delta t_{n+1,i} = \exp{ (\mu_{n+1,i} ) }$ for each node $v_i \in \cup e_{n+1}^r$.  For evaluating the performance, we find the Mean Absolute Error (MAE) between true values $t^{true}$ and estimated value $\hat{t}_{n,i}$ as, 
$
    \text{MAE} =  \frac{1}{N} \sum_{n=1}^{N}  \biggl[\sum_{v_i \in \cup e_n^r } \frac{1}{ | \cup e_n^r | } | t^{\text{true}}_n - \hat{t}_{n,i} | \biggr].
$
Additionally, the performance of predicting projected adjacency vectors, $ \bfa^r_{n, i}, \bfa^\ell_{n, i}  $, at time $t$ is calculated by the proportion of true neighbors in the estimated adjacency vectors, $ \hat{\bfa}^{r}_{n, i}, \hat{\bfa}^{\ell}_{n, i} $ as,
$
        \text{Recall} = \frac{1}{2N} \sum_{n=1}^{N}  \biggl[\sum_{v_i \in \cup e_n^r } \frac{1}{ | \cup e_n^r | } \frac{ { \left(\bfa^r_{n, i} \right) }^\intercal  \hat{\bfa}^{r}_{n, i} }{ {\left( {  \bfa^{r}}_{n, i} \right)}^{\intercal} {  \bfa}^{r}_{n, i} } \biggr] +  \frac{1}{2N} \sum_{n=1}^{N}  \biggl[\sum_{v_i \in \cup e_n^r } \frac{1}{ | \cup e_n^r | } \frac{ { \left(\bfa^\ell_{n, i} \right) }^\intercal  \hat{\bfa}^{\ell}_{n, i} }{ {\left( {  \bfa^{\ell}}_{n, i} \right)}^{\intercal} {  \bfa}^{\ell}_{n, i} } \biggr].
$

In all our experiments, we use the first $50\%$ of hyperedges for training, $25\%$ for validation, and the remaining $25\%$ for testing. We use the Adam optimizer, and models are trained for $100$ epochs with the batch size set to $128$ and a learning rate of $0.001$. Models have the same representation size of $d=64$, and we use $20$ negative hyperedges for each true hyperedge in the dataset. All the reported scores are the average of ten randomized runs, along with their standard deviation.
%

\subsection{Results}
Our proposed model, DHyperNodeTPP is evaluated against previous works and an undirected baseline model, HyperNodeTPP. The results in Table \ref{tab:dynamicdirectedlinkprediction} demonstrate that our models, HyperNodeTPP and DHyperNodeTPP, outperformed previous works, HGDHE and HGBDHE, in event time prediction. This superior performance can be attributed to the fact that our models are trained to predict the next event on the nodes, as opposed to previous models that are trained to predict the next event on a hyperedge, which is more difficult and prone to error as they have to predict events of longer duration. However, in our case, events on nodes are more frequent and have shorter periods, hence the smaller error in the event time prediction. 
%

Further, one can observe that the temporal models perform much better than the static model GAT. This shows the need for temporal models for real-world relation forecasting. In our work, we achieve this using a memory based temporal node representation learning technique as explained in Section \ref{sec:dynamic_node_embedding}. We can also observe that models that use attention-based temporal representation, HyperNodeTPP, and DHyperNodeTPP, perform better than non-attentional models, HGBDHE and HGDHE, by comparing the performance in event type prediction tasks. This justifies using neighborhood features for temporal node representation learning. Additionally, we have discussed the limitations of this model in 
Appendix \ref{sec:limiations}. We have also included scalability experiments and a comparison against the state-of-the-art event forecasting model in Appendix \ref{sec:appendix:scalable_training} and \ref{sec:appendix:comparing_with_state_of_the_art_event_prediction_model}, respectively.

\begin{figure*}[!htbp]
  \centering
  \begin{subfigure}{0.23\textwidth}
    \includegraphics[width=\textwidth]{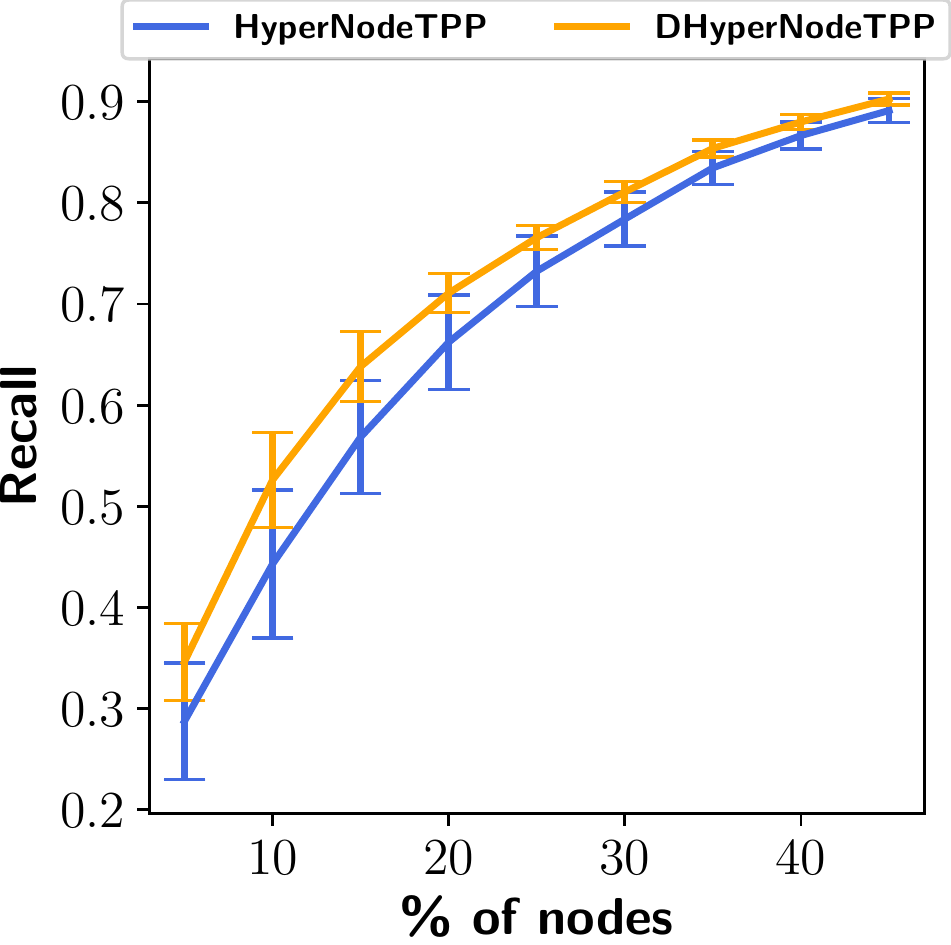}
    \caption{Adjacency forecasting}
    \label{fig:enron_connectivity}
  \end{subfigure}
  \begin{subfigure}{0.23\textwidth}
    \includegraphics[width=\textwidth]{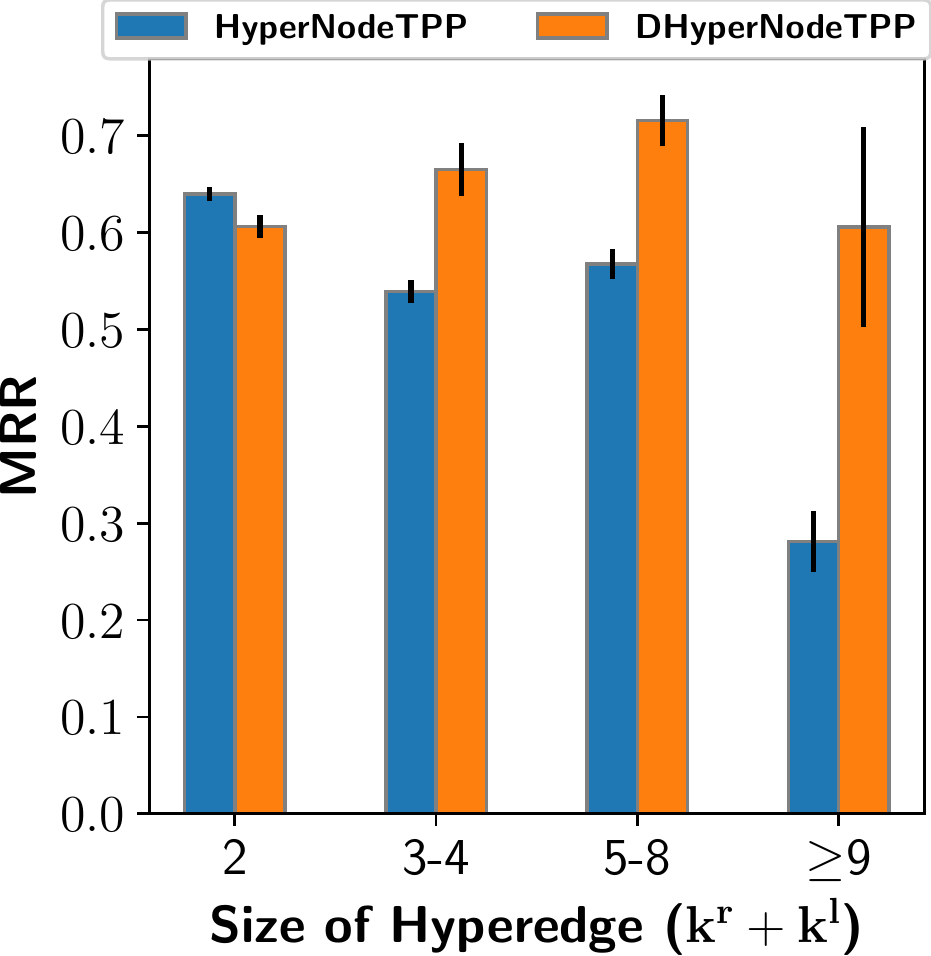}
    \caption{Event type prediction}
    \label{fig:enron_mrr}
  \end{subfigure}
  \begin{subfigure}{0.23\textwidth}
    \includegraphics[width=\textwidth]{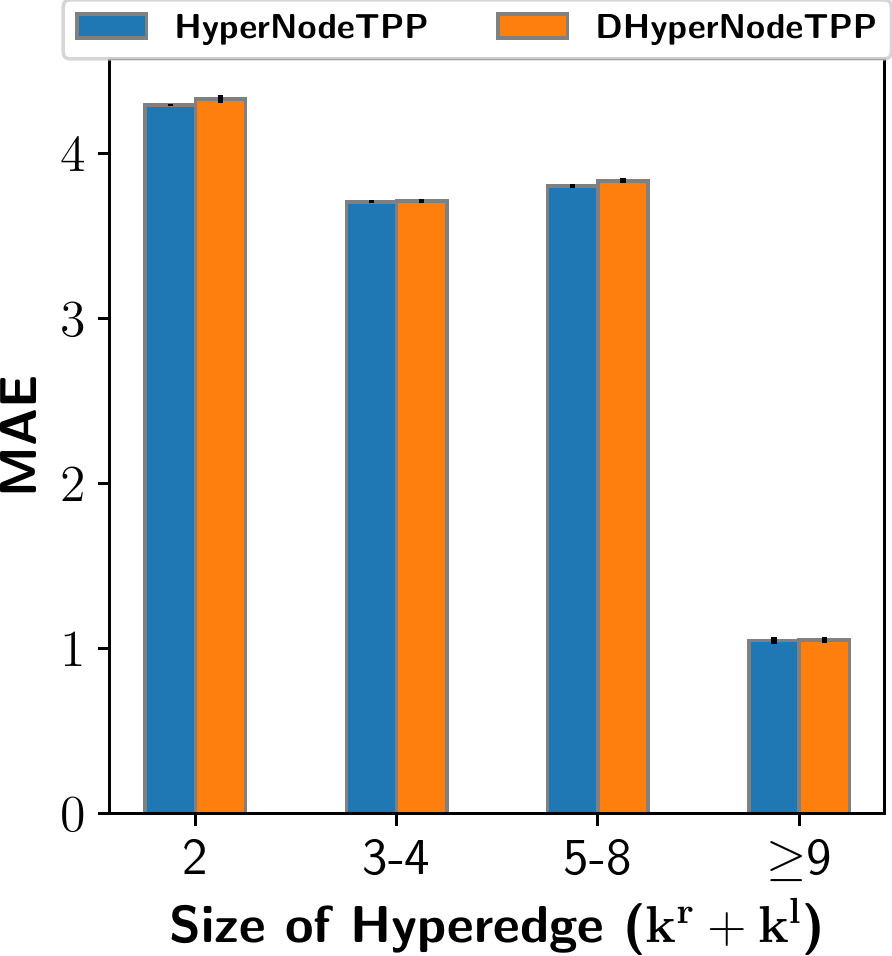}
    \caption{Event time prediction}
    \label{fig:enron_mae}
  \end{subfigure}
    \begin{subfigure}{0.23\textwidth}
    \includegraphics[width=\textwidth]{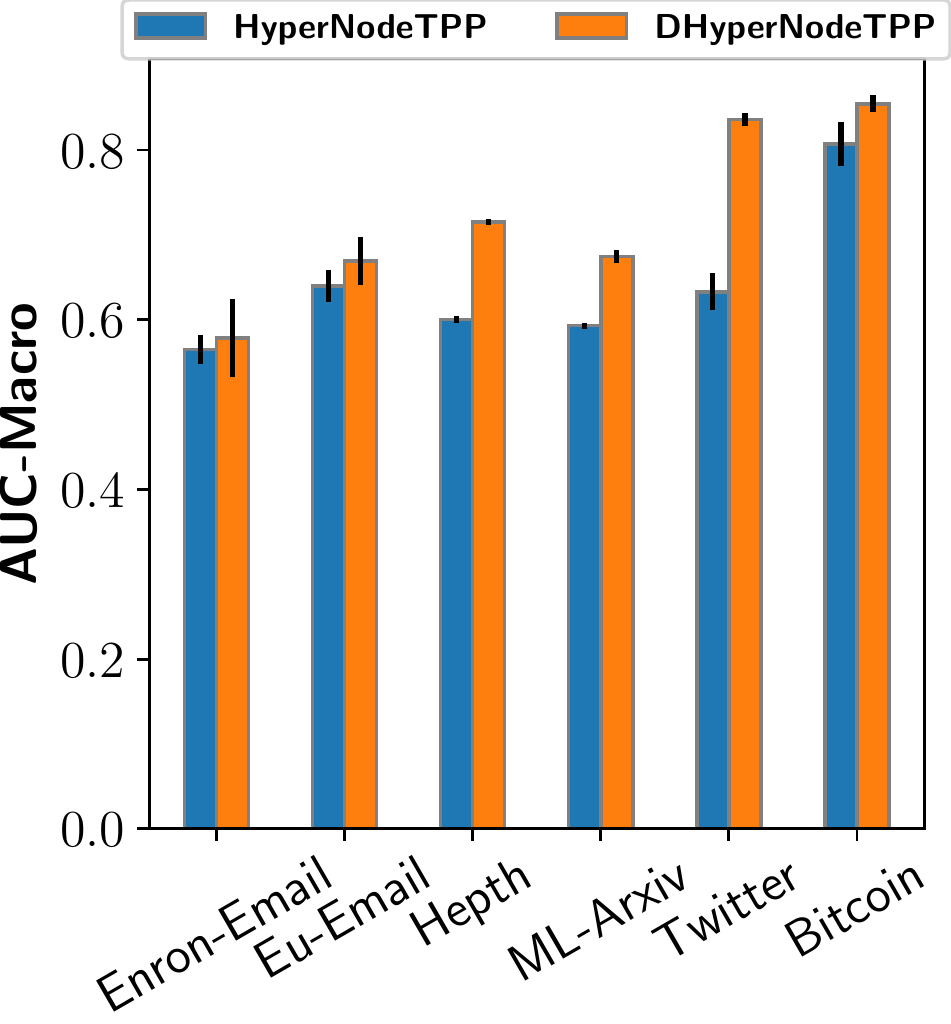}
    \caption{Hyperedge size prediction}
    \label{fig:size_macro}
  \end{subfigure}
  \caption{Comparison of the performance of our directed and undirected model on different forecasting tasks. Figures \ref{fig:enron_connectivity}, \ref{fig:enron_mrr}, \ref{fig:enron_mae} are on Enron-Email dataset. Here, we can observe that representation from DHyperNodeTPP performs better than HyperNodeTPP for adjacency forecasting. Furthermore, DhyperNodeTPP performs better than HyperNodeTPP for hyperedge sizes greater than two for the event type prediction. For event time prediction, both models perform equally, as events are modeled on nodes, and for hyperedge size prediction, directed models perform better. Hence, we can learn better representation using direction information.}
  \label{fig:multi_task_enron_performance}
\end{figure*}

\paragraph{Comparing directed and undirected models.} To observe the advantage of directed modeling, we compare the directed model DHyperNodeTPP with the undirected model HyperNodeTPP.  Here, we see an average increase of $70\%$ in the MRR metric for event type prediction in citation network based datasets Hepth and ML-Arxiv. The proposed model gives comparable performance to the undirected model for the other datasets. This is because in these datasets, around $70\%$ of relations have only a single node in the left hyperedge, and HyperNodeTPP is performing well on these hyperedges of size two ($k^r+k^l=2$). 
Figure \ref{fig:enron_mrr} shows for the Enron-Email dataset, HyperNodeTPP performs better than DHyperNodeTPP for hyperedges of size two, while for hyperedges of size greater than two, DhyperNodeTPP outperforms HyperNodeTPP. We have observed similar trends in Eu-Email, Twitter, and Bitcoin datasets, as shown in Figures 7b, 8b, and 11b. For Hepth and ML-Arxiv, the proposed directed model outperforms the undirected model in all hyperedge size groups, as shown in Figures 9b and 10b, respectively.
For the task of event time prediction, both models performed similarly in all datasets, as observed in Figure \ref{fig:enron_mae} and Appendix \ref{appendix:effect_of_size_on_performance}. This is because events are modeled on the nodes, not on the hyperedges. We also compared the models in the task of predicting projected adjacency vectors in the candidate hyperedge generation module. This is done by finding estimates for projected adjacency vectors by thresholding the estimated probability vectors, $\hat{\bfa}^r_{n,i} = \bbI_{\sigma( \bmtheta^r_{n,i} ) > thres } ,  \hat{\bfa}^\ell_{n,i}= \bbI_{\sigma( \bmtheta^\ell_{n,i} ) > thres }$, and recall is calculated. Here, $thres$ is selected so that a fixed percentage of nodes will be presented as neighbors. Figure \ref{fig:enron_connectivity} shows recall calculated for Enron-Email dataset with respect to the percentage of nodes by varying it from $5\%$ to $50\%$ in the step of $5\%$. Here, we can see DHyperNodeTPP has more recall than HyperNodeTPP, especially when the percentage of allowed neighbors is small and similar trends are observed in other datasets in Appendix \ref{appendix:adjacency_vector_prediction}. Further, in the case of hyperedge size prediction, DHyperNode outperforms HyperNodeTPP in all the datasets, as shown in Figure \ref{fig:size_macro}. The multi-label classification metric AUC-marco \citep{Fawcett:2006:AnIntroductionToROCAnalysis} is calculated to evaluate this task. Here, we can see our model DHyperNodeTPP outperforms HyperNodeTPP considerably. There is an average improvement of $13.3\%$ AUC macro metric.
 
%
%
%
\begin{table}
\centering
\small 
\begin{tabular}{lccc}
\toprule
\textbf{Methods}  & \textbf{HyperSAGNN}  & \textbf{CATSETMAT} & \textbf{Ours}  \\ 
\midrule
\textbf{iAF1260b}  & 60.1 $\pm$ 2.4  & 35.0 $\pm$ 1.4 & $\bm{61.3 \pm 0.5}$  \\
\textbf{iJO1366}   & 57.9 $\pm$ 1.8 & 36.2 $\pm$ 0.9 & $\bm{59.0 \pm 1.8}$ \\
\textbf{USPTO} &  35.9 $\pm$ 1.8 & 37.8 $\pm$ 0.5 & $\bm{38.1 \pm 0.6}$ \\
\bottomrule
\end{tabular}
\caption{Performance of our hyperedge predictor in Section \ref{sec:architecture_of_link_predictor} compared to previous works on static directed hyperedge prediction. Here, the MRR metric is used to evaluate the performance. We can observe that our proposed hyperedge predictor performs better than previous models.}
\label{tab:staticdirectedlinkprediction}
\end{table}
\paragraph{Comparing hyperedge models with the pairwise models.}  The advantage of hyperedge models over pairwise edge models can be inferred by comparing models DHyperNodeTPP to TGN, which is the pairwise equivalent of DHyperNodeTPP. Our model DHyperNodeTPP has an average improvement of $31.8\%$ in MRR metric over TGN in event type prediction. 
%
%
%
%

\paragraph{Comparing different hyperedge prediction architectures.} In Table \ref{tab:staticdirectedlinkprediction}, we compared our directed hyperedge predictor to previous works, HyperSAGNN~\citep{ZhangEtAL:2019:Hyper-SAGNNASelfAttentionBasedGraphNeuralNetworkForHypergraphs} for undirected hyperedge prediction, and CATSETMAT~\citep{SharmaEtAL:2021:CATSET} for bipartite hyperedge prediction. Our method performs considerably better than previous architectures. We get an average improvement of  $3.7\%$ over the undirected model and $46.6\%$ over the bipartite model in the MRR metric. This is because our link predictor models self-connections in both right and left hyperedges and cross edges between them.  The poor performance of CATSETMAT is because it focuses on modeling the cross connections and fails to model the self connections in both right and left hyperedge due to the bipartite assumption. 


\section{Conclusion}
\label{sec:conclusion}
In this work, we have proposed a comprehensive solution for the problem of learning representations for higher-order, directed, and temporal relations by presenting the model \textbf{DHyperNodeTPP}. Most of the previous works either reduce higher-order relations to pairwise edges~\citep{XuEtAL:2020:InductiveRepresentationLearningOnTemporalGraphs,TrivediEtAL:2019:DyRepLearningRepresentationsoverDynamicGraphs,KumarEtAL:2019:PredictingDynamicEmbeddingTrajectoryInTemporalInteractionNetworks,GraciousEtAL:2021:NeuralLatentSpaceModelForDynamicNetworksAndTemporalKnowledgeGraphs}
or ignore the directionality information in the relations~\citep{GraciousEtAL:2023:DynamicRepresentationLearningWithTemporalPointProcessesForHigherOrderInteractionForecasting}.  
So, to address this, we propose a sequential generative process involving three stages. The first stage is used to forecast nodes where the events will occur, which are then consumed by the second task to forecast candidate hyperedges occurring at the event time. The second task is then used by the directed hyperedge prediction stage to filter out the true hyperedges. Since the number of nodes that observe an event at a particular time is very small, we could considerably reduce the search space for the possible hyperedge. Further, we provide a temporal node representation learning technique that can do batch training by using a Memory Module. This reduces computational complexity while training on very large datasets. We also provide an architecture for directed hyperedge prediction by combining three levels of information in the hyperedges.  This work also involves creating five datasets.  We have shown the advantage of the proposed model over existing models for undirected and bipartite hyperedge forecasting models. Some interesting directions for future works are using advanced negative sampling techniques \citep{HwangEtAL:2022:AHPLearningToNegativeSampleForHyperedgePrediction} to make the model learn efficiently and provide good candidate hyperedges in a scalable way. 

\section*{Acknowledgements}
The authors would like to thank the SERB, Department of Science, and
Technology, Government of India, for the generous funding towards this
work through the IMPRINT Project: IMP/2019/000383.

\bibliography{references_thesis}
\appendix

\twocolumn[
\begin{center}
\textbf{\Large Neural Temporal Point Processes for Forecasting Directional Relations
in Evolving Hypergraphs: Appendix }
\end{center}
 \hfill \break
 \hfill \break
 ]

\noindent
 The notations used, and their definitions are shown in Tables \ref{tab: Notations}, \ref{tab: Notations for hyperedge predictor}, and \ref{tab: Notations_2}. The code and datasets used in our work can be found here,

\begin{table*}
\centering
    \begin{tabular}{|c|c|}
    \toprule
    \textbf{Notations}  &  \textbf{Definitions}  \\
    \midrule 
    $\calV$ & Set of nodes \\
    $\calH$ & Set of hyperedges \\
    $\calG = ( \calV, \calH )$ &  Hypergraph \\ 
    $\bbI_{(\cdot)}$ & Indicator function. $\mathbb{I}_{p}$ returns one if condition $p$ is true, else zero. \\
    $\bbR$ & Real space\\
    ${}^{n}\calP_{r}$ & Permutations\\
    $v_i$ & $i$th node \\
    $h_i$ & $i$th hyperedge/higher-order relation \\
    $h_i^r, h_i^\ell$ & Left and right hyperedges \\
    $k^r$, $k^ell$ & Sizes of left and right hyperedges \\ 
    $\lambda(t)$ & Conditional intensity function\\
    $t$ & Time \\
    $L_n$ &  Concurrent higher-order relations occurring at time $t_n$ \\
    $\calE (t)$ & $ \calE (t)=\{(e_{1}, t_{1}), \ldots, (e_n, t_n)\}$ is the history till time $t$\\
    $\bfV (t) \in \bbR^{\calV \times d}$  & Node representations \\
    $\bbR$ & Real space \\ 
    $m$ & Concurrent hyperedge index \\ 
    $n$ & Index of the event \\
    $k$ & Combined Hyperedge size ($k = k^r + k^\ell$) \\
    $\rmP( )$ & Probability \\ 
    $\rmP^*_i()$, $\rmS^*_i ()$ &  Conditional Probability and Survival function for node $v_i$\\
    $\mu_i(t_{n+1})$ & Parameters of the time modeling for node $v_i$\\
    $\MLP_t (. )$ & MLP for time modeling \\ 
    $s_t$ & Variance of $\mathrm{Lognormal}$ distribution \\ 
    $ \MLP_{ar} (. ) , \MLP_{a \ell} ( .) $ & MLP for adjacency modeling \\ 
    $  \MLP_{sr} ( .), \MLP_{s \ell} (.) $ & MLP for size modeling \\
    $\calO (.)$ & Order of complexity \\ 
    $N$ & Total number of events in the dataset \\
    $e_n$ & $L_n$ Concurrent events at time $t_n$\\
    $L_n$ & Number of concurrent events in $e_n$\\
    $\Phi(.)$ & Cumulative density function of standard normal \\
    $\sigma $ & Sigmoid layer \\
    $i$ & Index of node \\
    $d$ & Size of the node representation \\ 
    $\cup $ & Union \\ 
    $\bfA_{n}^r,\bfA_{n}^\ell$ &  Right and left projected adjacency matrix for hyperedges in $e_{n}$ \\ 
    $\bfK^r_{n} , \bfK^\ell_{n} $ &  Right and left size vectors for hyperedges in $e_{n}$ \\ 
    $\sigma (. )$ & Sigmoid layer \\ 
    $\calL \calL_t^n$ & Negative log-likelihood for time modeling at time step $t_{n}$ \\
    $\calL \calL_{s}^{n}$ & Negative log-likelihood for size modeling at time step $t_{n}$\\ 
    $\calL \calL_{a}^{n}$ & Negative log-likelihood for projected adjacency matrix modeling at time step $t_{n}$\\
    $\calL \calL_{h}^{n}$ & Negative log-likelihood for hyperedge prediction at time step $t_{n}$\\ 
    $\bmkappa^r_{n,i}, \bmkappa^\ell_{n,i}$  & Parameters of right and left size distribution at time step $t_n$ \\ 
    $\bmtheta_{n,i}^r , \bmtheta_{n,i}^r $ &  Parameters of right and left adjacency matrices at time step $t_n$\\ 
    $\calL \calL$ & Complete negative log-likelihood \\ 
    $\calH^{c}_{n}$ & Candidate hyperedges at time $t_{n}$ \\ 
    $\lambda_{ h}(t)$ & Hyperedge link predictor \\ 
    $\bfW_{t1}, \bfW_{t0}, b_{t1}, \bfb_{t0}$ & Parameters of event modeling task\\
    $\bfW_{sr1}, \bfW_{sr0}, \bfb_{sr1}, \bfb_{sr0}$ & Parameters of right hyperedge size prediction model \\
    $\bfW_{ar0} , \bfb_{ar0}$ & Parameters of right projected adjacency vector prediction model \\ 
    \bottomrule
    \end{tabular}
\caption{Notations used till Section \ref{sec:architecture_of_link_predictor}}
\label{tab: Notations}
\end{table*}

\begin{table*}
\centering

    \begin{tabular}{|c|c|}
    \toprule
    \textbf{Notations}  &  \textbf{Definitions}  \\
    \midrule 
    $ \bfW_{{CQ}^{r}}, \bfW_{{CK}^{r}}, \bfW_{{CV}^{r}}$ & Parameters of $\mathrm{CAT}$ layer for the right hyperedge  \\ 
    $ \bfW_{{CQ}^{\ell}}, \bfW_{{CK}^{\ell}}, \bfW_{{CV}^{\ell}} $ & Parameters of $\mathrm{CAT}$ layer for the left hyperedge \\ 
    $\bfW_{SQ}, \bfW_{SK}, \bfW_{SV}$ & Parameters of $\mathrm{SAT}$ layer \\ 
    $\bfW_{o^r}, b_{o^r}$ & Final layer parameters for the right hyperedge in the directed hyperedge  predictor \\ 
    $v_{i^{r}}, v_{i^{\ell}}$ & Node in the right hyperedge and left hyperedge \\
    $o_{i^r}$ &  Final output score for a node $v_{i^{r}}$ in the right hyperedge  \\
    $\calP^{h^r}, \calP^{h^\ell}$ & Pooled output scores for the right and left hyperedges \\ 
    $\bfd_{i^r}^{ch}  $ & Dynamic representation from cross-attention for the node  $v_{i^{r}}$  in right hyperedge\\
    $\bfd_{i^r}^{sh}  $ & Dynamic representation from self-attention for the node  $v_{i^{r}}$ in right hyperedge\\
    $\bfd_{i^r}^h $ & Dynamic representation of the node  $v_{i^{r}}$ in right hyperedge \\
    $e_{ij}$ & Attention inner-product score \\
    $\alpha_{ij}$ & Attention weights \\
    $\bfW_{s^r}$ & Static hyperedge representation  linear layer for right hyperedge \\
    $\bfs_{i^r}^{h}$ & Static hyperedge representation for  the node  $v_{i^{r}}$  right hyperedge\\
    $f(\  )$ & Scoring function for conditional intensity \\
    \bottomrule
    \end{tabular}
\caption{Notations for the Directed Hyperedge Predictor in Section \ref{sec:architecture_of_link_predictor}}
\label{tab: Notations for hyperedge predictor}
\end{table*}

\begin{table*}
\centering
    \begin{tabular}{|c|c|}
    \toprule
    \textbf{Notations}  &  \textbf{Definitions}  \\
    \midrule 
    $c$ & Event type \\
    $\mathcal{C}$ & Category \\ 
    $\mathcal{T}$ & Historical event times \\
    $\ud \tau$ & Derivative of time $\tau$ \\
    $t^p_{i}$ & Previous event time for node $v_i$ \\
    $ \bfW^s, \bfW^r,   \bfW^\ell, \bfb_v$ & Learnable parameters for dynamic node representation \\
    $\mathbf{Mem}_{i}$ & Memory entries for node $v_i$ \\ 
    $\mathbf{Mem}$ & Memory Module \\ 
    $b$ & Batch size \\ 
    $\lambda_h(t)$ & Conditional intensity function of hyperedge $h$\\
    $t^p_h$ & Previous time of event time for hyperedge $h$ \\
    $\rmP (t)$/ $\rmP_h(t)$ & Probability of the event $h$ \\
    $\rmS (t)$ & Survival function for the event \\ 
    $\bmpsi (t)$ & Function to calculate Fourier features for time $t$\\
     $\{\omega_i \}_{i=1}^d$ and  $\{\phi_i \}_{i=1}^d$  & Learnable parameters of $\bmpsi(t)$  \\
     $\calB$ & Batch size \\ 
     $\mathbf{msg}_{i}^{r} (t_{b} )$ & Messages vector created for node $v_i$ in the right hyperedge at time $t_b$ \\
     $\calN$ & Hyperparameter for the neighborhood features \\
     $\mathcal{N}_{h^r}(t)$ & Recent $\calN$ relations involving node $v_i$ in the right hyperedge\\
     $\mathcal{N}_{h^\ell}(t)$ & Recent $\calN$ relations involving node $v_i$ in the left hyperedge\\ 
     $\mathrm{MultiHeadAttention}$ & Multi-head attention \\
     $\bfq(t)$ & Query vector \\
     $\bfC(t) $ & Key vector \\
     $\bfC(t)$ & Value vector \\ 
     $\hat{t}_{n,i}$ & Next event time estimate for $n$th event \\ 
     $\hat{\bfa}^r_{n,i}, \hat{\bfa}^\ell_{n,i}$ & Projected adjacency vector estimates for $n$th event \\ 
    \bottomrule
    \end{tabular}
    \caption{Notations used after Section \ref{sec:architecture_of_link_predictor}}
\label{tab: Notations_2}
\end{table*}

\section{Background on Temporal Point Process}
A temporal point process (TPP) defines a probability distribution over discrete events happening in continuous time. It can be shown as a sequence of timestamps $\calE(t_n) = \{t_0, t_1, \ldots, t_n\}$, here $t_n$ is the time of occurrence of the event. The goal of the temporal point process is to model the probability of the next event given the history $\rmP(t_{n+1}| t_0, \ldots, t_{n} )$.  A marked temporal point process is a type of TPP, where each event is associated with a category $c \in \mathcal{C}$. Here, $\calC$ can be a set of discrete objects or a feature vector of real values. This can be represented as $\calE(t_n) = \{(t_0,c_0), (t_1,c_1), \ldots, (t_n,c_n)\}$. A temporal point process can be parameterized by an intensity function $\lambda(t)$,
\begin{align}
    \lambda(t) \ud t = \rmP(\text{event in $[t, t+ \ud t]$ }\:|\: \calE (t_n) ). 
\end{align}
Then, the likelihood  of the next event time can be modeled as 
    \begin{align} \label{eq: event probability}
        \rmP(t|  \calE (t_n) ) & = \lambda(t) \rmS(t) \nonumber , \\
       \rmS (t) &  =  \exp\left ( \int_{t_n}^t -\lambda (\tau) \ud  \tau ) \right) .
    \end{align}
Here,  $\rmS (t)$ is the survival function representing the probability of the event not occurring during the interval $[t_n, t)$. The likelihood for the entire observation $\calE (T) $ in the duration $[0, T]$ can be written as follows, 
    \begin{align}
     \rmP ( \calE (T)) &= \prod_{n=0}^{|\calE (t)|} \lambda(t_n) \rmS (T). 
    \end{align}
Here,  $\rmS (T) = \prod_{n=0}^{|\calE (t)|} \rmS (t_n) = \exp \left( \int_{0}^T -\lambda (\tau) \ud \tau \right) $.

The popular functional forms of a temporal point process are exponential distribution $\lambda(t) = \mu$, where $\mu > 0$ is the rate of event. Hawkes process \citep{HawkesEtAL:1973:SepctraofSomeSelf-ExcitingAndMutuallyExcitingPointProcesses} is another widely used method when events have self-exciting nature with $\lambda(t) = \mu + \alpha \sum_{t_n \in \calE (t)} \calK (t -t_n) $. Here, $\mu \geq 0$ is the base rate, $\calK (t) \geq 0 $ is the excitation kernel, and $\alpha \geq 0 $ is the strength of excitation. If $\lambda(t) = f(\calE (t)$ with $f(.)$ as a neural network, then it is called a neural temporal point process \citep{MeiEtAL:2017:TheNeuralHawkesProcessANeurallySelfModulatingMultivariatePointProcess}.
\section{Loglikelihood Derivations} \label{sec:appendix_loglikelihood_derivations}

\section{MLP Layers} \label{sec:mlp_layers_appendix}
\begin{itemize}
    \item $\mathrm{MLP}_t (\cdot ) $  parameterizes the mean of the $\mathrm{Lognormal}$ distribution in the event modeling task using the dynamic node representation $\bfv_i (t_n)$ as shown below,
    \begin{align} 
        \mu_{n+1, i} = & \MLP_t (\bfv_i(t_n) ) \nonumber \\ 
        =  & \bfW_{t1} \mathrm{tanh}( \bfW_{t0} \bfv_i (t_n) + \bfb_{t0} ) + b_{t1}.
    \end{align}
    Here, $\bfW_{t1} \in \bbR^{1 \times d}, \bfW_{t0} \in \bbR^{d \times d}, b_{t1} \in \bbR, \bfb_{t0} \in \bbR^{d}$.
    
    \item $\mathrm{MLP}_{sr} (\cdot )$  parameterizes the $\mathrm{Bernoulli}$ distributions of the sizes of right hyperedges for the nodes in right hyperedges as shown below,
    \begin{align}
        \bmkappa^r_{n+1,i} = &\MLP_{sr}(\bfv_i (t_n)) \nonumber \\
        = & \bfW_{sr1} \mathrm{tanh}( \bfW_{sr0} \bfv_i (t_n) + \bfb_{sr0} ) + \bfb_{sr1}, \nonumber\\
        & \forall\: v_i \in \cup e_n^r.
    \end{align}
    Here, $\bfW_{sr1} \in \bbR^{ k^r_{max} \times d}, \bfW_{sr0} \in \bbR^{d \times d}, \bfb_{sr1} \in \bbR^{k^r_{max}}, \bfb_{sr0} \in \bbR^{d  }$. Similarly, we model $\mathrm{MLP}_{sl} (\cdot ) $ for the sizes of left hyperedges with respect to nodes in the right. 
    
    \item $\mathrm{MLP}_{ar} (\cdot )$ parameterizes the $\mathrm{Bernoulli}$ distributions of the projected adjacency vectors for the nodes in the right hyperedge as shown below,
    \begin{align}
    \bmtheta^r_{n+1,i} &= \MLP_{ar}(\bfv_i (t_n)) \nonumber \\
    &=  \mathbf{Mem}\: \mathrm{tanh}( \bfW_{ar0} \bfv_i (t_n) + \bfb_{ar0} ), \nonumber \\ & \forall\: v_i \in \cup e_n^r .
    \end{align}
    Here, $  \mathbf{Mem} \in \bbR^{|\calV| \times d },\bfW_{ar0}  \in \bbR^{d \times d }, \bfb_{ar0} \in \bbR^d$. The final layer is parameterized by the memory defined in Section \ref{sec:dynamic_node_embedding} to capture the history of events on nodes. Similarly, we model $\MLP_{al} (\cdot)$ to parameterize the distribution of left projected adjacency vectors.
\end{itemize}

\section{Hyperedge Predictor Architecture Details}
\label{sec:attention_layers_appendix}
 
\begin{figure}
    \centering
    \includegraphics[width=0.98\linewidth]{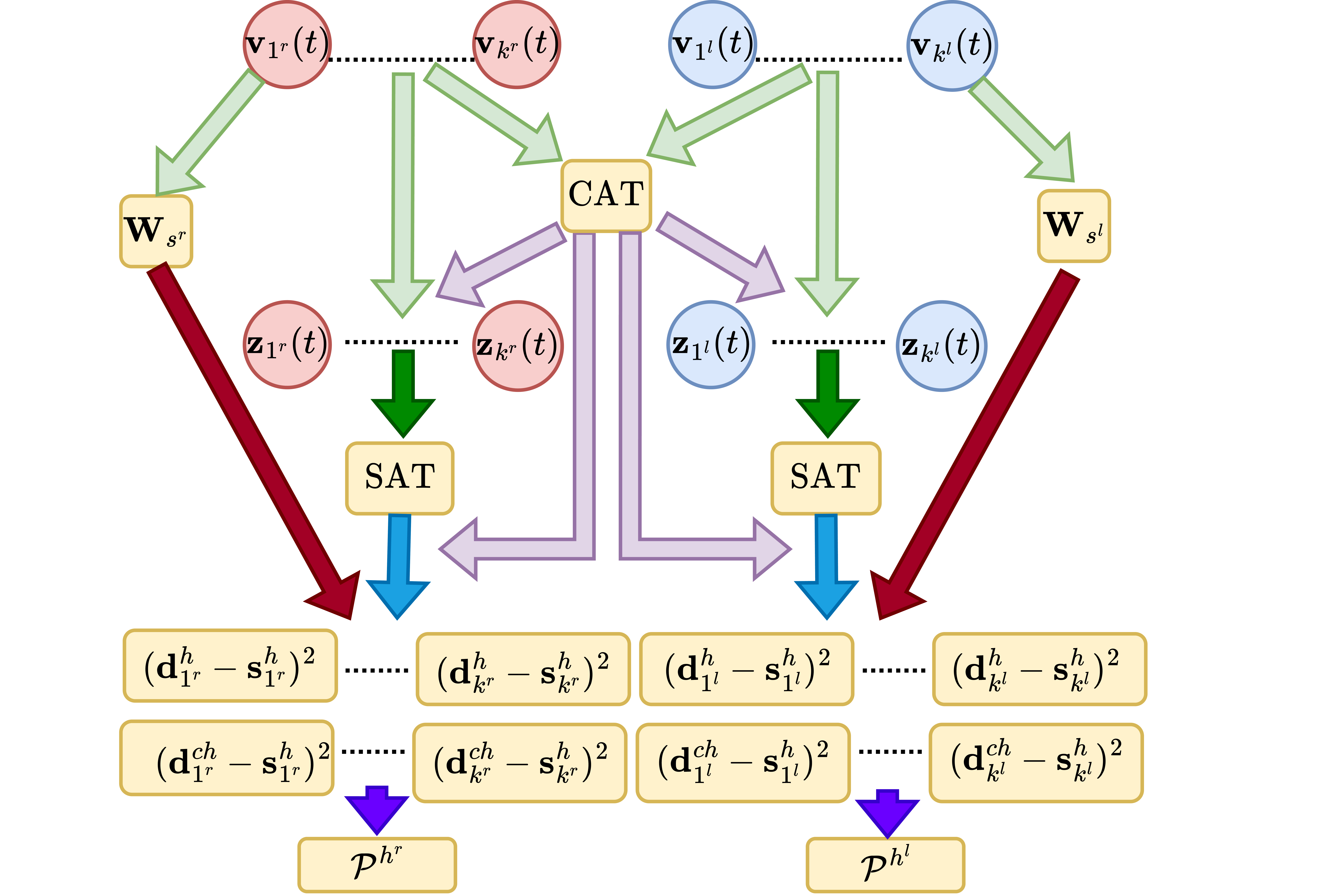}
    \caption{Hyperedge predictor architecture}
    \label{fig:link_prediction}
\end{figure}

Figure \ref{fig:link_prediction} shows the block diagram of the directed hyperedge predictor used in our model DHyperNodeTPP. The following are the attention layers used in our model.
\paragraph{CAT}  Given a directed hyperedge $h=(\{v_{1^r}, \ldots, v_{k^r} \}, \{ v_{1^{\ell}}, \ldots, v_{k^{\ell}}  \} )$, we use the cross-attention layer between the sets of nodes to create dynamic representation as shown below, 
    \begin{align}
          e_{i^{r}j^{\ell}}  &= ( \bfW_{CQ^r}^{\intercal} \bfv_i (t))^\intercal \bfW_{{CK}^{\ell}}^{T} \bfv_{j^{\ell}} (t),\nonumber \\
          & \text{$  1^r \leq i^r \leq k^r, 1^\ell \leq  j^{\ell} \leq k^{\ell} $}, \nonumber \\ 
          e_{i^{\ell}j^r}  &= ( \bfW_{{CQ}^{\ell}}^{\intercal} \bfv_{i^{\ell}} (t))^\intercal \bfW_{CK^r}^{T}  \bfv_{j^r} (t), \nonumber \\
          & \text{$ 1\leq i^{'} \leq k^{'}, 1^r \leq  j^r \leq k^r $} \nonumber \\
          \alpha_{{i^r} j^{\ell}} &= \frac{\displaystyle \exp{ (e_{ {i^r} j^{\ell}}}) }{\displaystyle \sum_{ 1 \leq \ell^{\ell} \leq k^{\ell} } \exp{ (e_{i^r \ell^{\ell}}}) } ,
             \alpha_{i^{\ell}j^r} = \frac{\displaystyle \exp{ (e_{i^{\ell}j^r}}) }{\displaystyle \sum_{ 1 \leq \ell^r \leq k^r } \exp{ (e_{i^{\ell}\ell^r}}) }.
    \end{align}
These weights are used to find the cross dynamic hyperedge representation for node $v_{i^{r}}$ in the right hyperedge and node $v_{i^{\ell}}$ in the left hyperedge as follows, 
    \begin{align} \label{eq:cross_dynamic_embeddings}
        \bfd^{ch} _{i^r}  &= \tanh\left( \sum_{1^\ell \leq j^{\ell} \leq k^{\ell}} \alpha_{i^r j^{\ell}} \bfW_{ {CV}^{\ell}}^{\intercal} \bfv_{j^{\ell}}(t) \right) \nonumber \\ 
        \bfd^{ch}_{i^{\ell}}  &= \tanh\left( \sum_{1^r \leq j^r \leq k^r} \alpha_{i^{\ell}j^\ell} \bfW_{CV^r}^{\intercal} \bfv_{j^r}(t) \right) .
    \end{align}
Here, $ \bfW_{CQ^r}, \bfW_{CK^r}, \bfW_{CV^r} \in \bbR^{d \times d} $ are the learnable weights for the right hyperedge, and $ \bfW_{{CQ}^{\ell}}, \bfW_{{CK}^{\ell}}, \bfW_{{CV}^{\ell}} \in \bbR^{d \times d} $ are the learnable weights left hyperedge.


\begin{table*}
    \centering
    \scalebox{0.8}{%
    \begin{tabular}{lccccccccc}
        \toprule
            \textbf{Datasets}  &    $|\mathcal{V}|$  & $|\mathcal{E}(T)|$ & $|\mathcal{H}^r|$ & $|\mathcal{H}^{\ell}|$  & $T$ & $\Delta t (mean)$ &  $\Delta t (max)$ &  $\Delta t (min)$ & $L (max)$\\
            \midrule 
            \textbf{Enron-Email} & 183  & 10,311 & 1,003 & 89 & 99,070.18 & 6.67 & 1,589.89 & 0.0008 & 3\\
            \textbf{Eu-Email} &  800  & 208,403 &  11,897 & 744 & 69,459,254 & 333.294 & 23,484,865 & 1 & 3 \\ 
            \midrule
            \textbf{Twitter} & 2,130 & 9,889 & 1,218 & 2,321 & 17,277.80 & 1.877 & 41.39 & 0.18 & 3 \\ 
            \midrule 
            \textbf{Hepth} & 451 & 9,882 & 8,384 & 1,352 & 21,532.70 & 2.17 & 275.90 & 0.0001 & 1\\
            \textbf{ML-Arxiv} & 659 & 18,558  & 2,995 & 17,014 & 62,741.56  &  3.38 & 314.62 & 0 & 6 \\
            \textbf{Bitcoin} &  999 & 2,31,071 & 23,901 & 6,706 & 1,416,317,420 & 62.89 & 5,408.00 & 0 & 6 \\
            \bottomrule
    \end{tabular}}
     \caption{Datasets used for Temporal Directed Hypergraphs and their vital statistics.}
    \label{tab:datasets_temporal}
\end{table*}
\begin{table*}
    \centering
    \begin{tabular}{lccccccccc}
        \toprule
            \textbf{Datasets}  & $Datatype $  & $|\mathcal{V}|$  & $|\mathcal{E}|$ & $|\mathcal{H}^r|$ & $|\mathcal{H}^{\ell}|$\\
            \midrule 
            \textbf{iAF1260b} & metabolic reactions & 1668  & 2084 & 2010 & 1985\\
            \midrule 
            \textbf{iJO1366} & metabolic reactions &  1805  & 2253 &  2174 & 2146 \\ 
            \midrule
            \textbf{USPTO} & organic reactions & 16293 & 11433 & 6819 & 6784\\
        \bottomrule
    \end{tabular}
     \caption{Datasets used for static Directed Hypergraphs and their vital statistics.}
    \label{tab:datasets_static}
\end{table*}


\paragraph{SAT} Given a set of vectors $(\{\bfz_{1}, \ldots, \bfz_{k} \}$, with $\bfz_{i} \in \bbR^d$. The self-attention based dynamic representations are calculated as follows, 
 \begin{align}
        e_{ij} &= ( \bfW_{SQ}^{\intercal} \bfz_i (t))^\intercal \bfW_{SK}^{\intercal} \bfz_j (t), \text{$\forall  1 \leq i,j \leq k, i \neq j $}, \nonumber \\
        \alpha_{ij} &= \frac{\displaystyle \exp{ (e_{ij}}) }{\displaystyle \sum_{ 1 \leq \ell \leq k, i \neq \ell } \exp{ (e_{i\ell}}) }.
    \end{align}
These weights are used to calculate the second dynamic hyperedge representation for each node $v_i$ as, 
    \begin{align} \label{eq:self_dynamic_embeddings}
        \bfd^{sh}_i  = \tanh\left( \sum_{1 \leq j \leq k, i \neq j } \alpha_{ij} \bfW_{SV}^{\intercal} \bm{z}_j  (t) \right) .
    \end{align}
Here, $\bfW_{SQ}, \bfW_{SK}, \bfW_{SV} \in \bbR^{d \times d}$ are learnable weights. 

\section{Dataset Descriptions} 
\label{sec:datasets_description_appendix}

\subsection{Temporal Datasets}
\label{sec:datasets_temporal_appendix}
Table \ref{tab:datasets_temporal} shows the statistics of the temporal datasets. Here, $|\mathcal{V}|$ denotes the number of nodes, $|\mathcal{E}(T)|$ denotes the number of hyperedges, $|\mathcal{H}^{r}|$ denotes the number of unique right hyperedges, $|\mathcal{H}^{\ell}|$ denotes the number of unique left hyperedges, $T$ is the time span of the dataset, and $\Delta t (mean), \Delta t (max), \Delta t (min)$, are the mean, maximum and minimum values of inter-event duration, defined as the time difference between two consecutive events. $T, \Delta t (mean), \Delta t (max)$, and $\Delta t (min)$ are calculated after scaling the original time by the mean value of interevent duration. $L (max)$ denotes the maximum number of concurrent hyperedges. The following are the temporal-directed hypergraph datasets used to train the model.

\paragraph{Enron-Email.} \footnote{http://www.cs.cmu.edu/~enron/} This dataset comprises email exchanges between the employees of Enron Corporation. Email addresses are the nodes in the hypergraph. A temporal directed hyperedge is created by using the sender's address as the left hyperedge, the recipients' addresses as the right hyperedge, and the time of exchange as the temporal feature. Further, the hyperedge size is restricted to 25, the less frequent nodes are filtered out, and only exchanges involving Enron employees are included.

\paragraph{Eu-Email.} \citep{YinEtAL:2017:LocalHigherOrderGraphClustering} This is an email exchange dataset between members of a European research institution and the temporal directed hyperedge is defined exactly to that of the Enron-Email dataset. The hyperedge size is restricted to 25, and the less frequent nodes were filtered out.
 
\paragraph{Twitter.} \citep{ChodrowEtAL:2019:AnnotatedHypergraphsModelsAndApplications} This dataset contains tweets about the aviation industry exchanged between users for 24 hours. Here, directed hyperedge is formed by senders and receivers of a tweet.
\paragraph{Hepth.} \citep{GehrkeEtAL:2003:OverviewOfThe2003KDDCup} This is a citation network from ArXiv, initially released as a part of KDD Cup 2003. It covers papers from January 1993 to April 2003 under the Hepth (High Energy Physics Theory) section. The author IDs are the nodes, the directed hyperedge is between the authors of the paper and the cited papers, and the publication time is taken as the time of occurrence. The hyperedge size is restricted to 25, and less frequent nodes are filtered out.

\paragraph{ML-Arxiv.} This is a citation network created from papers under the Machine Learning categories - "cs.LG," "stat.ML," and "cs.AI" in ArXiv\footnote{https://arxiv.org/help/bulk\_data}. The definition of the hyperedge is identical to that of the Hepth dataset. All papers published before 2011 are filtered out, and nodes are restricted to authors who have published more than 20 papers during this period. Citations between these papers are considered to create a directed hyperedge.

\paragraph{Bitcoin.}  \citep{WuEtAL:2022:DetectingMixingServicesViaMiningBitcoinTransactionNetworkWithHybridMotifs} This is a set of bitcoin transactions that happened in the month of November 2014. Here, each user account is a node, and the transactions are represented as a directed hyperedge where coins are debited from one group of users and credited to another group.  Here, we removed transactions where the left and right groups have the same set of nodes. 

\subsection{Static Datasets}
\label{sec:datasets_static_appendix}
Table \ref{tab:datasets_static} shows the statistics of the static datasets. Here, $Datatype$ denotes the type of the data, $|\mathcal{V}|$ denotes the number of nodes, $|\mathcal{E}|$ denotes the number of hyperedges, $|\mathcal{H}^{r}|$ denotes the number of unique right hyperedges, and $|\mathcal{H}^{\ell}|$ denotes the number of unique left hyperedges. Following are the static directed hypergraph datasets used to test the performance of directed hyperedge link predictor in Section \ref{sec:architecture_of_link_predictor} against previous state-of-the-art architectures.
    \paragraph{iAF1260b.} This is a metabolic reaction dataset  from BiGG Models \footnote{http://bigg.ucsd.edu/}. It is a knowledge base of genome-scale metabolic network reconstructions. 
    \paragraph{iJO1366.} Similar to the above dataset, this is a metabolic reaction dataset from BiGG Models. 
    \paragraph{USPTO.} This dataset consists of reactions from USPTO-granted patents \citep{JinEtAL:2017:PredictingOrganicReationOutcomesWithWLNetwork}. This dataset has been processed and prepared to consist of organic reactions created using a subset of chemical substances containing only carbon, hydrogen, nitrogen, oxygen, phosphorous, and sulfur.

\section{Baseline Models}
\label{sec:baseline_models_appendix}

\subsection{Pairwise Models}
\paragraph{TGN.} Temporal Graph Networks (TGN) \citep{RossiEtAL:2020:TemporalGraphNetworksForDeepLearningOnDynamicGraphs}  is the state-of-the-art model for pairwise edge forecasting in temporal networks. Here, a memory module followed by an attention layer is used to learn temporal node representations~\citep{XuEtAL:2020:InductiveRepresentationLearningOnTemporalGraphs}. To adapt this model to predict higher-order relations,  we use clique expansion to convert hyperedge to pairwise edges. For example, given a directed hyperedge event $(\{v_{1^r}, \ldots, v_{k^r} \}, \{ v_{1^{\ell}}, \ldots, v_{k^{\ell}}  \} )$ at time $t$,  we create two different sets of pairwise edges using clique expansions; i) $\{ ((v_{1^r}, v_{2^r}),t ) , \ldots,  (v_{ {k-1}^{r}}, v_{k^r}),t )  \}$ using pairwise expansion of right hyperedge and ii) $\{ ((v_{1^r}, v_{2^{\ell}}),t ) , \ldots,  (v_{ {k-1}^{r} }, v_{k^{\ell}}),t )  \}$ using cross edges between left and right hyperedge. We use separate edge prediction models to estimate the probability of each type of edge.

\paragraph{GAT.} The Graph Attention Network (GAT) \citep{VeličkovićEtAL:2018:GraphAttentionNetworks} model uses the attention mechanism to give importance scores for different neighbors for each node, allowing it to focus on the most informative parts of the graph. To adapt GAT to hypergraphs, we used a similar approach as TGAT but removed the time stamp as GAT does not consider temporal features of the graph. For predicting an edge at time $t$,  we construct a pairwise graph with all edges that occurred before time $t$.

\subsection{Hyperedge Models}
\paragraph{HGDHE, HGBHDE.} \citep{GraciousEtAL:2023:DynamicRepresentationLearningWithTemporalPointProcessesForHigherOrderInteractionForecasting} These models use a temporal point process for modeling higher-order relations. Here, for each possible hyperedge, a TPP is defined with its intensity as a function of the temporal node representation, $$\lambda(h_{n+1,j }) = \mathrm{Softplus}( f( \{ \bm{v}_i (t)\}_{v_i \in  h_{n+1,j} } )).$$ So, the likelihood for event $(e_{n+1}, t_{n+1})$ given  $\mathcal{E}(t_n)$ can be written as,
\begin{align*}
    \rmP((e_{n+1}, t_{n+1}) | \mathcal{E}(t_n) ) &=  \prod_{h_{n+1,j} \in e_{n+1}}  \lambda_{h_{n+1,j}}(t_{n+1}) \nonumber \\
        &\exp \left(- \sum_{h \in \mathcal{H}} \int_{t_{n}}^{t_{n+1}} \lambda_{h}(\tau) \ud \tau ) \right) .
\end{align*}
Here, $\mathcal{H}$ is all possible hyperedges. The loss is calculated over all events as follows,
\begin{align}
    \calN \calL \calL = & \sum_{n=0}^{N-1}  \sum_{h_{n+1,m} \in e_{n+1}} \log \lambda_{h_{n+1,m}}(t_{n+1}) - \nonumber \\ & \sum_{h \in \mathcal{H}} \int_{t_{n}}^{t_{n+1}} \lambda_{h}(\tau) \ud \tau ) .
\end{align}
However, these models are not scalable as they use each hyperedge as a separate event type, and the number of possible hyperedges is exponential to the number of nodes, so these models cannot be applied to real-world networks. 
    
HGDHE is used for modeling undirected hyperedge with HyperSAGNN \citep{ZhangEtAL:2019:Hyper-SAGNNASelfAttentionBasedGraphNeuralNetworkForHypergraphs} decoder for $f(.)$ with temporal node representation learned as a function of time and historical relations.
When a node is involved in an event, its memory entries are updated based on the features of the event using a recurrent neural network architecture. 
The updated memory is used for temporal representation calculation with Hypergraph convolution layer \citep{BaiEtAL:2021:HypergraphConvolutionAndHypergraphAttention} and Fourier  \citep{XuEtAL:2020:InductiveRepresentationLearningOnTemporalGraphs} time features to model the evolution of the node during the interevent time.
HGBDHE is used for modeling bipartite hyperedges with CATSETMAT \citep{SharmaEtAL:2021:CATSET} decoder for hyperedge prediction, but here we use them for modeling directed hyperedges. Here, left and right hyperedge have different temporal node representations but follow the same architecture of HGDHE.

To predict the duration of future events for these models, we have to calculate the expected time $t$ with respect to the event probability distribution, as shown below, 
\begin{align} \label{eq:durationprediction}
    \hat{t} &= \int_{t^p_h}^{\infty} (\tau - t_n ) \rmP_h(\tau) \ud \tau \\ 
    \rmP_h(t) &=\lambda_{h}(t) \exp\left( -   \int_{t_h^p}^{\infty} \lambda_{h}(\tau) \ud \tau ) \right). 
\end{align}
We can forecast the type of event at time $t$ by finding the $h_i$ with the maximum conditional intensity value, 
$
    \hat{h} = \argmax_{h_i} \lambda_{h_i}(t) .
$

\begin{figure*}[htbp]
  \centering
  \begin{minipage}[b]{0.27\textwidth}
    \centering
    \includegraphics[width=\textwidth]{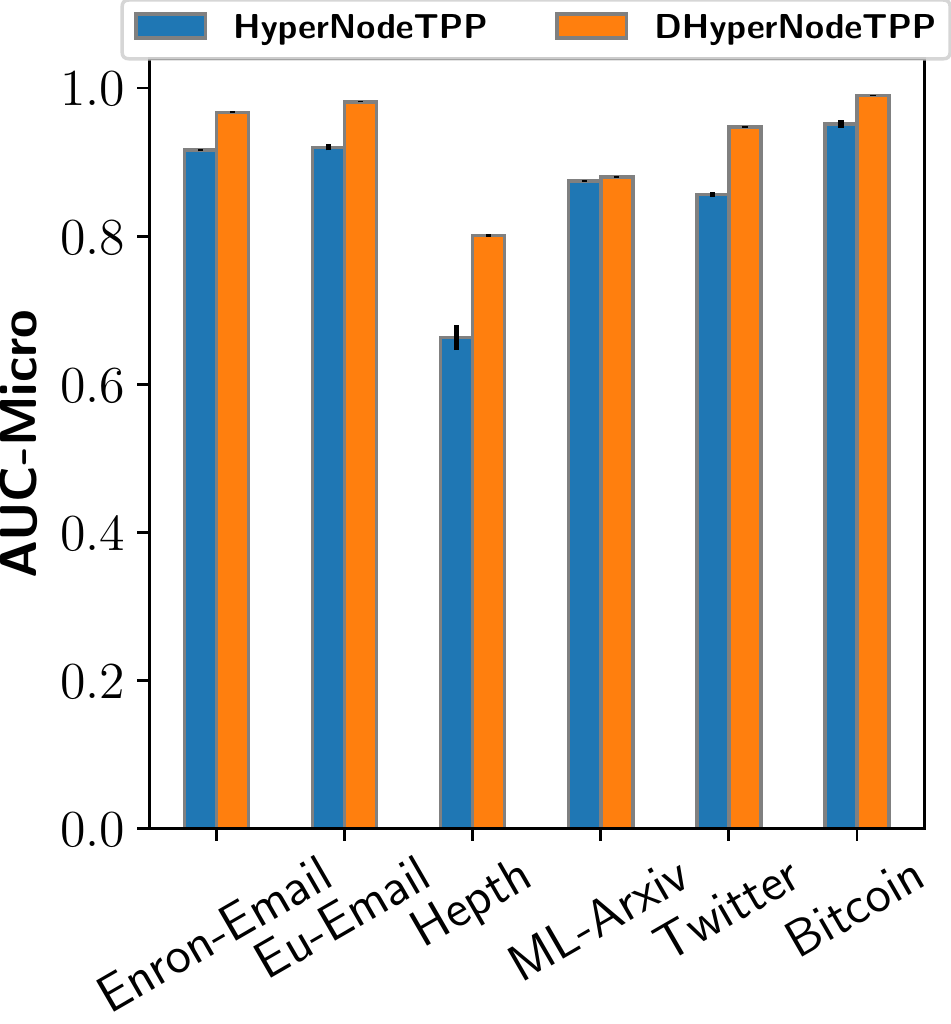}
    \caption{Hyperedge size prediction}
    \label{fig:size_prediction_auc_metrics_micro}
  \end{minipage}
  \hfill
  \begin{minipage}[b]{0.70\textwidth}
    \centering
    \small 
    \setlength{\tabcolsep}{2pt}
    \begin{tabular}{llcc}
    \toprule 
    \textbf{Datasets} &  Metrics & \textbf{HyperNodeTPP} & \textbf{DHyperNodeTPP} \\ 
    \midrule 
    \multirow{2}{*}{\textbf{Enron-Email}} & AUC macro & 56.4 $\pm$ 1.70 & $\bm{57.8 \pm 4.60}$ \\  
                                    & AUC micro & 91.5 $\pm$ 0.01 & $\bm{96.7 \pm 0.12}$ \\ 
    \multirow{2}{*}{\textbf{Eu-Email}  } & AUC macro & 63.9 $\pm$ 1.80 & $\bm{66.9 \pm 2.70}$ \\ 
                                        & AUC micro &  91.9 $\pm$ 0.40 & $\bm{98.1 \pm 0.05}$ \\
    \midrule
    \multirow{2}{*}{\textbf{Twitter}} &  AUC macro & 63.2 $\pm$ 1.80 & $\bm{83.6 \pm 0.07}$ \\ 
                                        & AUC micro &  85.6 $\pm$ 0.30 & $\bm{94.7 \pm 0.05}$ \\
    \midrule
    \multirow{2}{*}{\textbf{Hepth}} &  AUC macro & 60.0 $\pm$ 0.40 & $\bm{71.5 \pm 0.32}$ \\ 
                                        & AUC micro &  66.3 $\pm$ 1.70 & $\bm{80.0 \pm 0.17}$ \\
    \multirow{2}{*}{\textbf{ML-Arxiv}} &  AUC macro & 59.3 $\pm$ 0.35 & $\bm{67.4 \pm 0.77}$ \\ 
                                        & AUC micro &  87.4 $\pm$ 0.10 & $\bm{87.9 \pm 0.17}$ \\
    \midrule
     \multirow{2}{*}{\textbf{Bitcoin}} &  AUC macro & 80.71 $\pm$ 0.00 & $\bm{85.45 \pm 0.00}$ \\ 
                                        & AUC micro &  95.15 $\pm$ 0.00 & $\bm{98.97 \pm 0.00}$ \\
    \bottomrule
    \end{tabular}
    \captionof{table}{Hyperedge size prediction AUC scores in $\%$ \label{tab:size_prediction_auc_metrics} }
  \end{minipage}
\end{figure*}

\section{Ablation Studies}
\label{appendix:ablation_studies}

\begin{figure}[!htbp]
    \centering
    \includegraphics[width=0.85\linewidth]{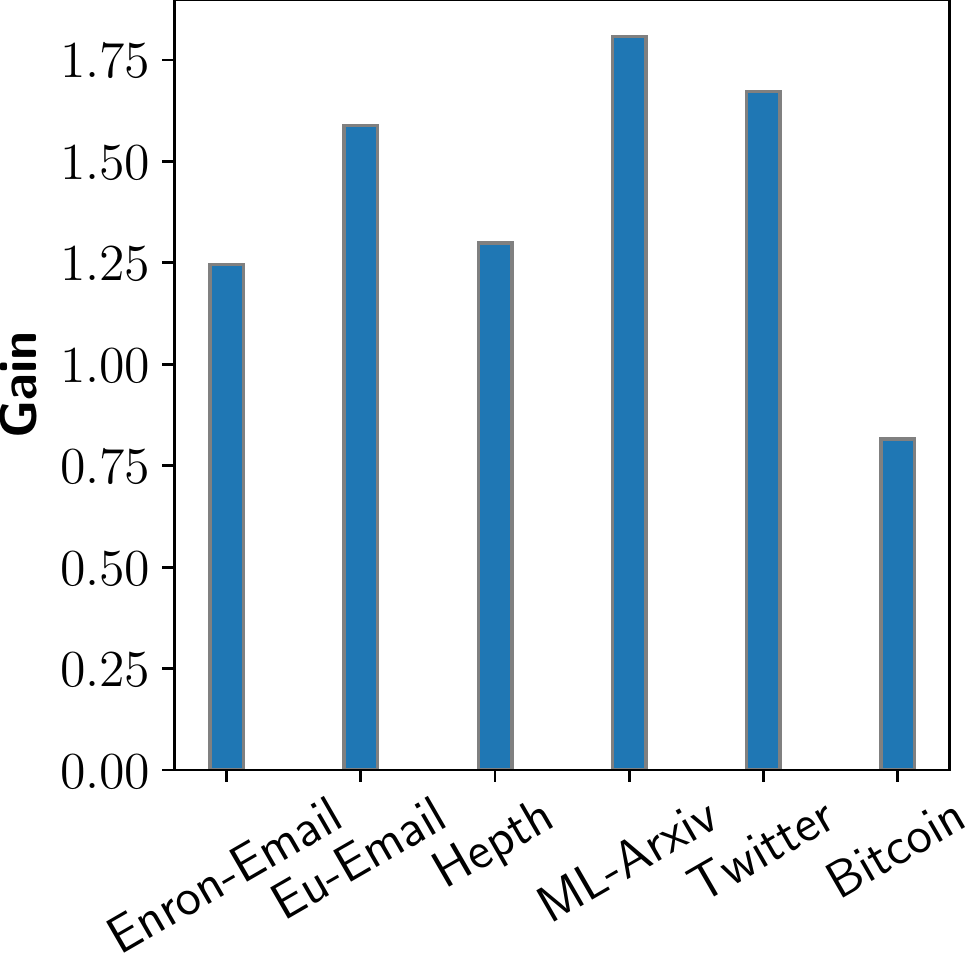}
    \caption{Gain in training speed obtained when batch size is increased to 128 from 32. Here, we can observe that increasing batch size reduced the computation time, except for Bitcoin dataset. Further, gains can be obtained by using parallel computation techniques.}
    \label{fig:gain_in_speed}
\end{figure}

\subsection{Effect of size on performance}
\label{appendix:effect_of_size_on_performance}
\begin{figure*}
  \centering
  \begin{subfigure}{0.31\textwidth}
    \includegraphics[width=0.95\textwidth]{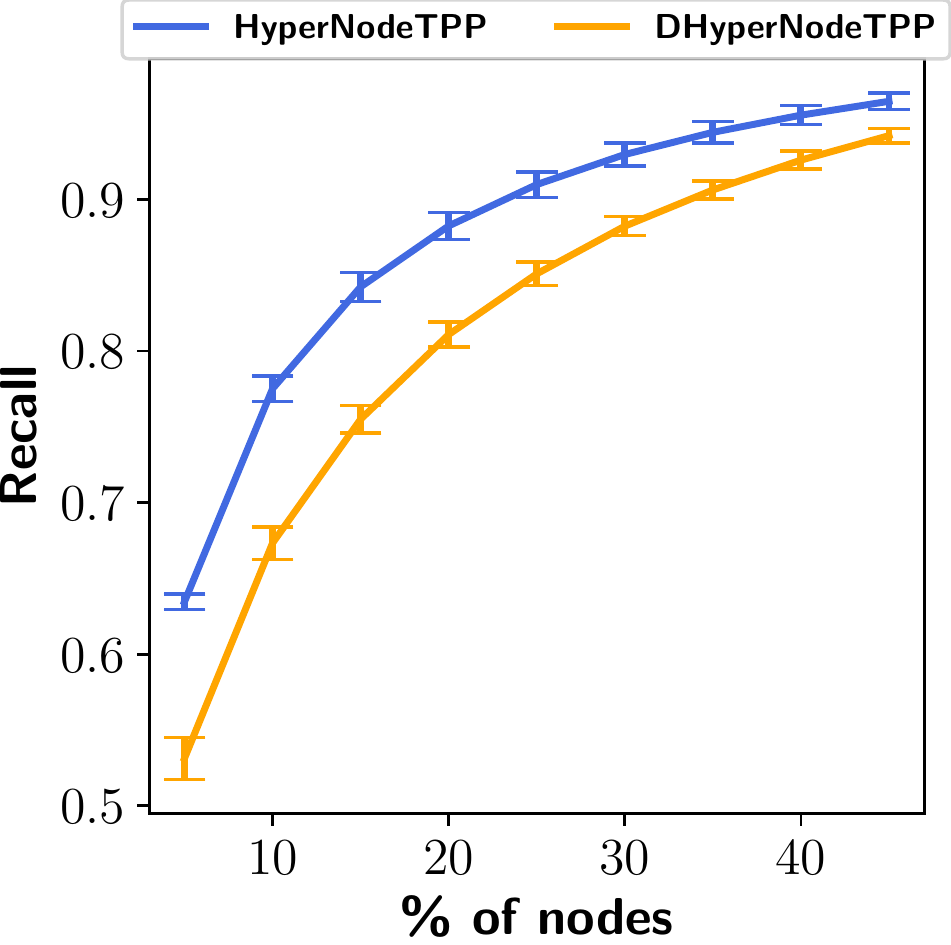}
    \caption{Adjacency forecasting}
    \label{fig:eumail_connectivity}
  \end{subfigure}
  \begin{subfigure}{0.31\textwidth}
    \includegraphics[width=0.95\textwidth]{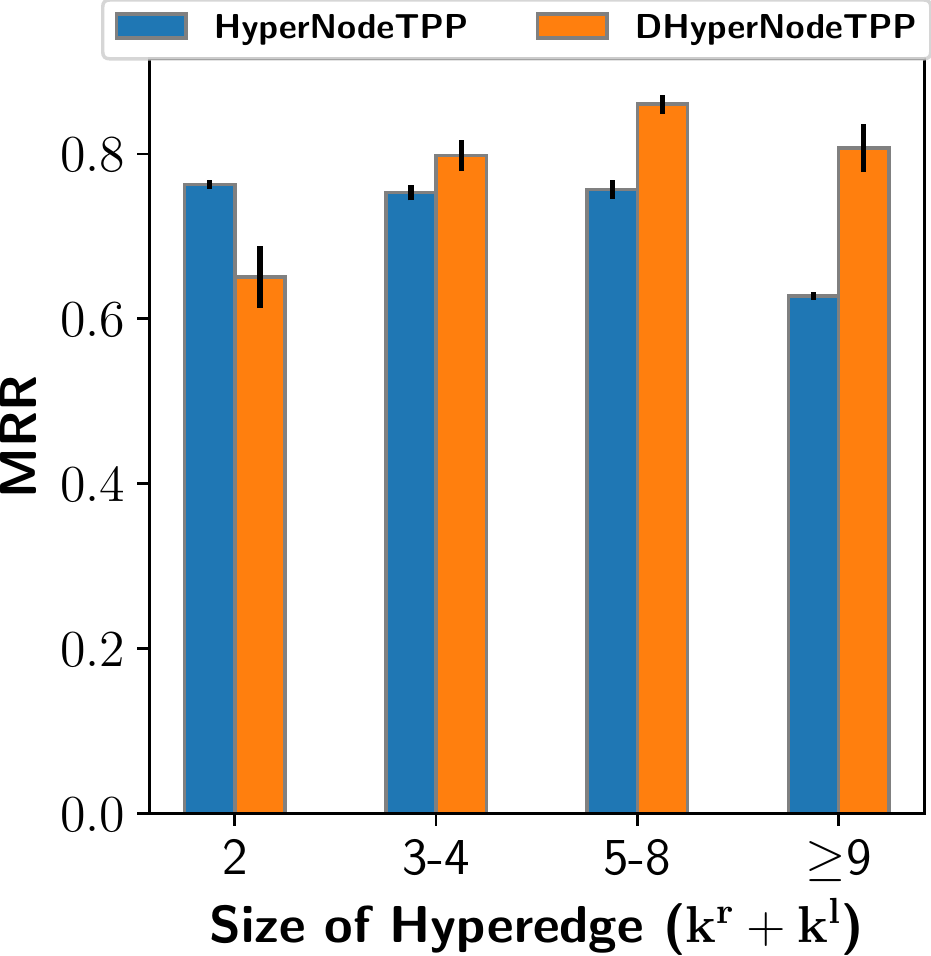}
    \caption{ Event type prediction }
    \label{fig:eumail_mrr}
  \end{subfigure}
  \begin{subfigure}{0.31\textwidth}
    \includegraphics[width=0.95\textwidth]{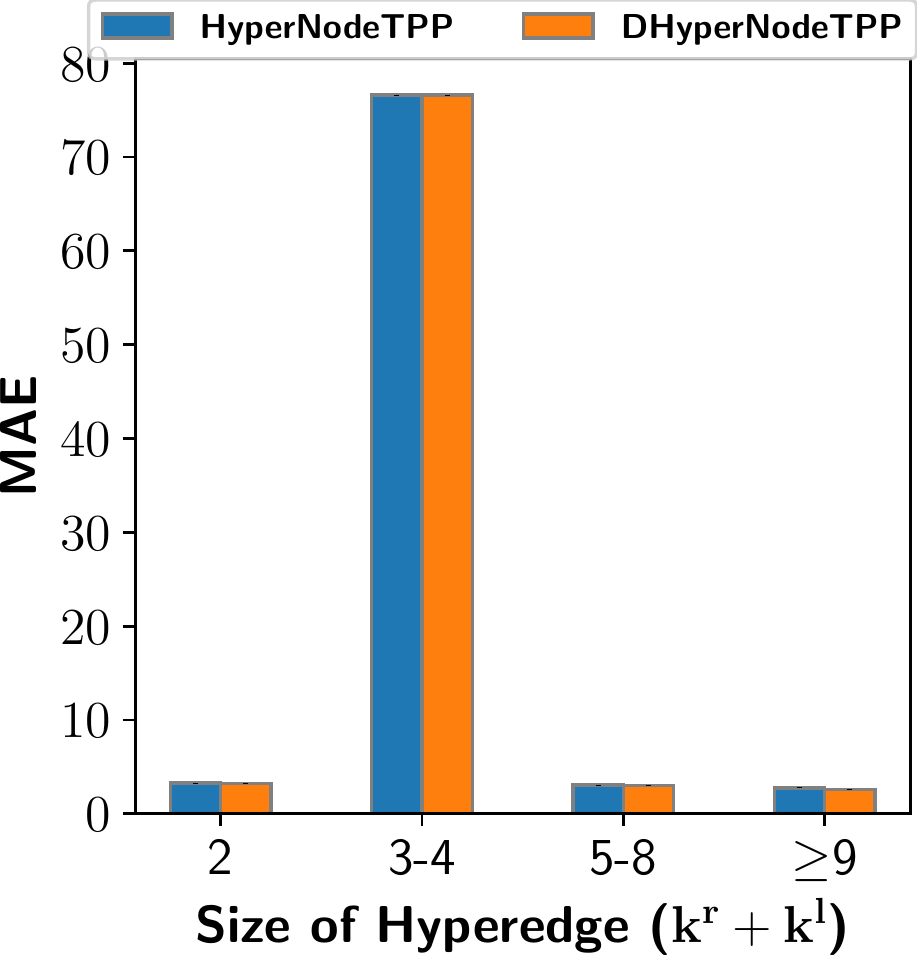}
    \caption{ Event time prediction}
    \label{fig:eumail_mae}
  \end{subfigure}
  \caption{Comparison of the performance of our directed and undirected model on different forecasting tasks on Eu-Email dataset. Here, we can observe that representation from HyperNodeTPP performs better than DHyperNodeTPP for adjacency forecasting. Further, HyperNodeTPP outperforms DHyperNodeTPP for event type prediction in hyperedges of size two. This is because, in  Eu-Email dataset, around $81\%$ percentage of hyperedges are of size two, and HyperNodeTPP overfits on these hyperedges. For event time prediction, both models perform equally, as events are modeled on nodes.}
  \label{fig:multi_task_eumail_performance}
\end{figure*}

\begin{figure*}
  \centering
  \begin{subfigure}{0.31\textwidth}
    \includegraphics[width=0.95\textwidth]{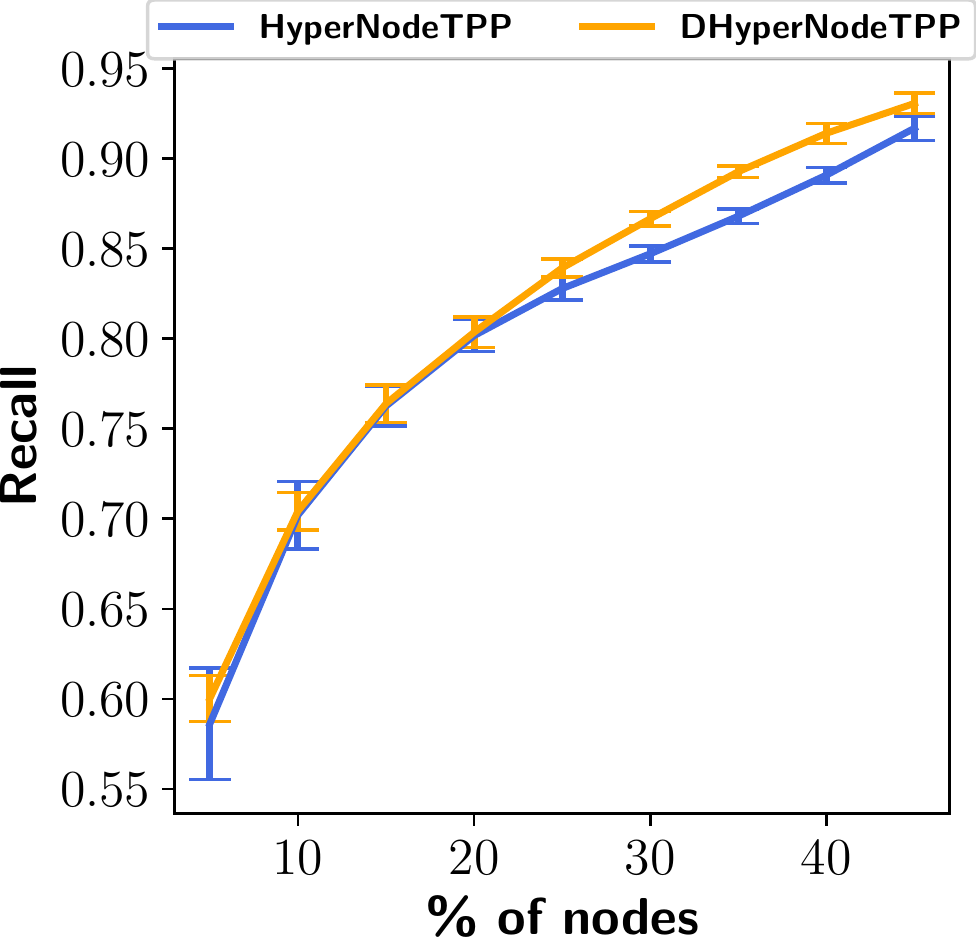}
    \caption{Adjacency forecasting}
    \label{fig:twitter_connectivity}
  \end{subfigure}
  \begin{subfigure}{0.31\textwidth}
    \includegraphics[width=0.95\textwidth]{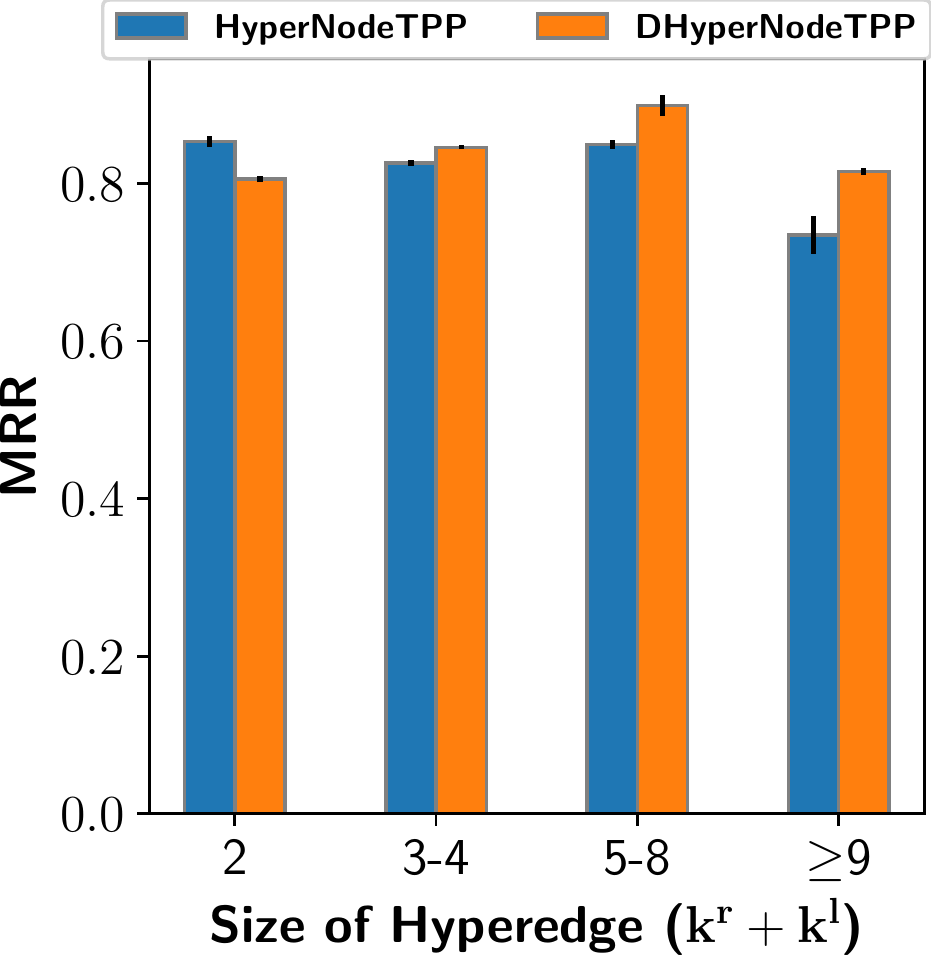}
    \caption{ Event type prediction}
    \label{fig:twitter_mrr}
  \end{subfigure}
  \begin{subfigure}{0.31\textwidth}
    \includegraphics[width=0.95\textwidth]{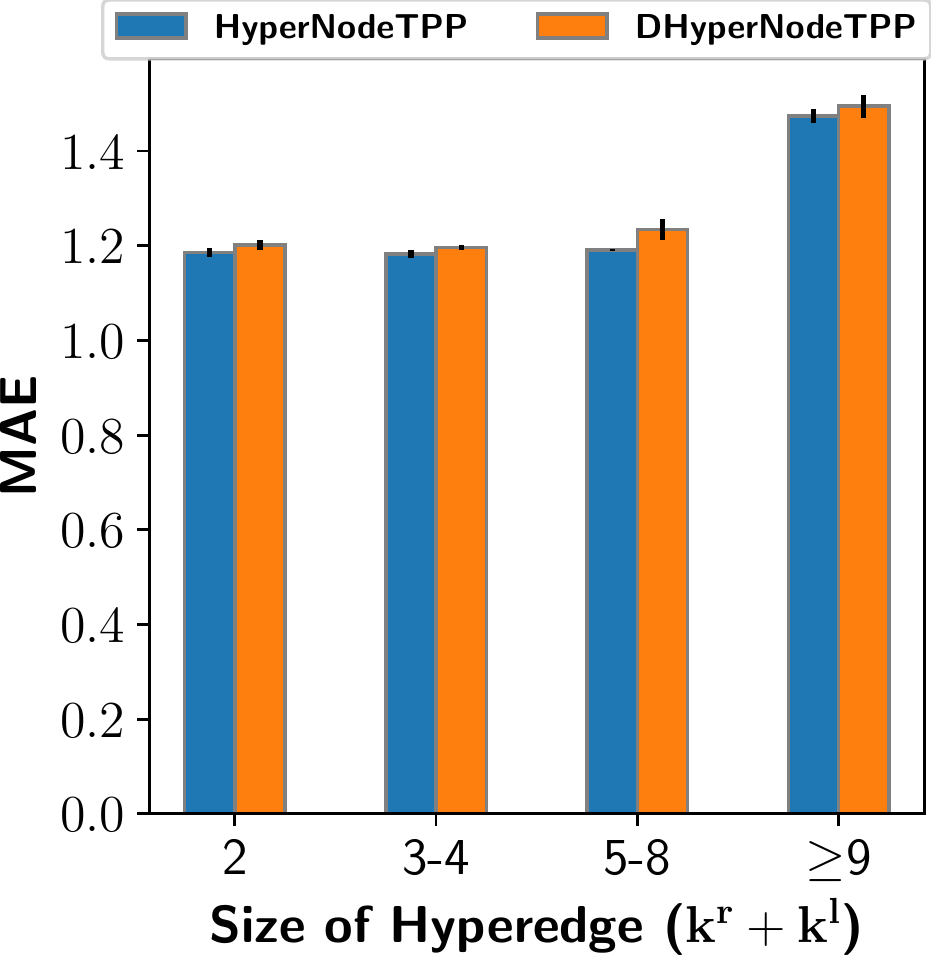}
    \caption{ Event time prediction}
    \label{fig:twitter_mae}
  \end{subfigure}
  \caption{Comparison of the performance of our directed and undirected model on different forecasting tasks on Twitter dataset. Here, we can observe that representation from DHyperNodeTPP performs better than HyperNodeTPP for adjacency forecasting. Furthermore, DhyperNodeTPP performs better than HyperNodeTPP for hyperedge sizes greater than two for the event type prediction. In the Twitter dataset, around $68\%$ hyperedges are of size two, and HyperNodeTPP overfits on these samples.
  For event time prediction, both models perform equally, as events are modeled on nodes, and for hyperedge size prediction, directed models perform better. Hence, we can learn better representation using direction information.}
  \label{fig:multi_task_twitter_performance}
\end{figure*}

\begin{figure*}
  \centering
  \begin{subfigure}{0.31\textwidth}
    \includegraphics[width=0.95\textwidth]{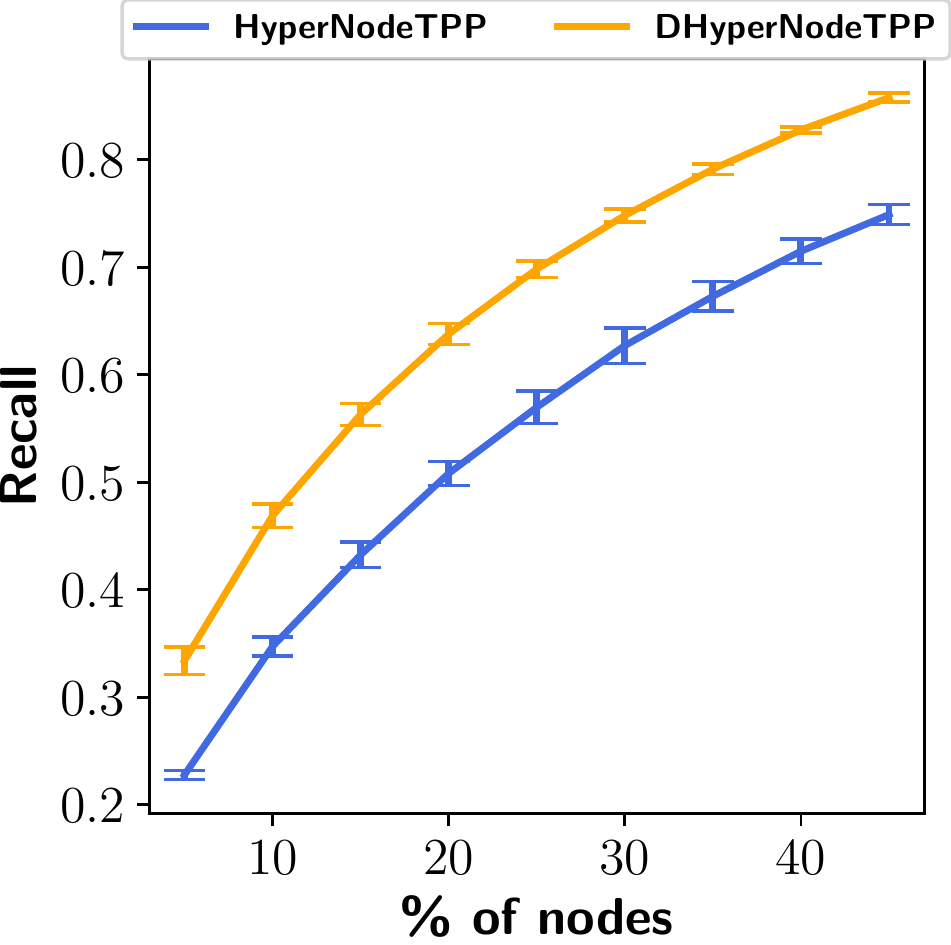}
    \caption{Adjacency forecasting}
    \label{fig:hepth_connectivity}
  \end{subfigure}
  \begin{subfigure}{0.31\textwidth}
    \includegraphics[width=0.95\textwidth]{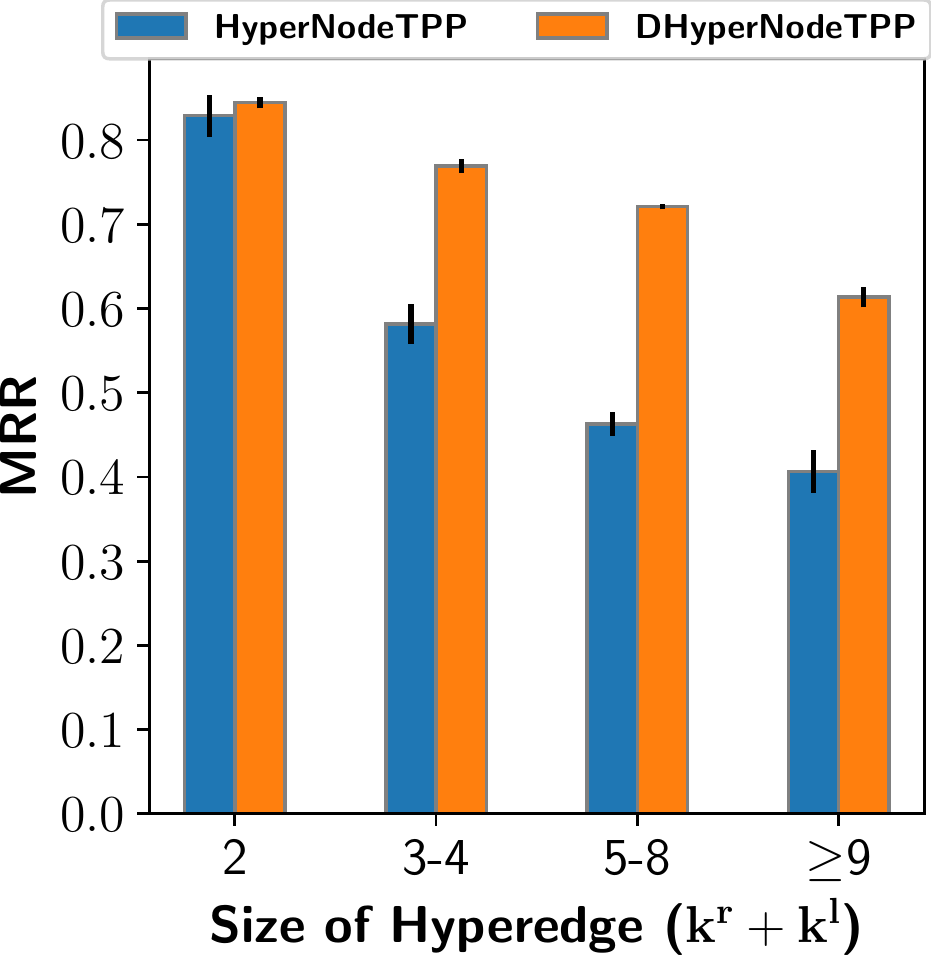}
    \caption{ Event type prediction}
    \label{fig:hepth_mrr}
  \end{subfigure}
  \begin{subfigure}{0.31\textwidth}
    \includegraphics[width=0.95\textwidth]{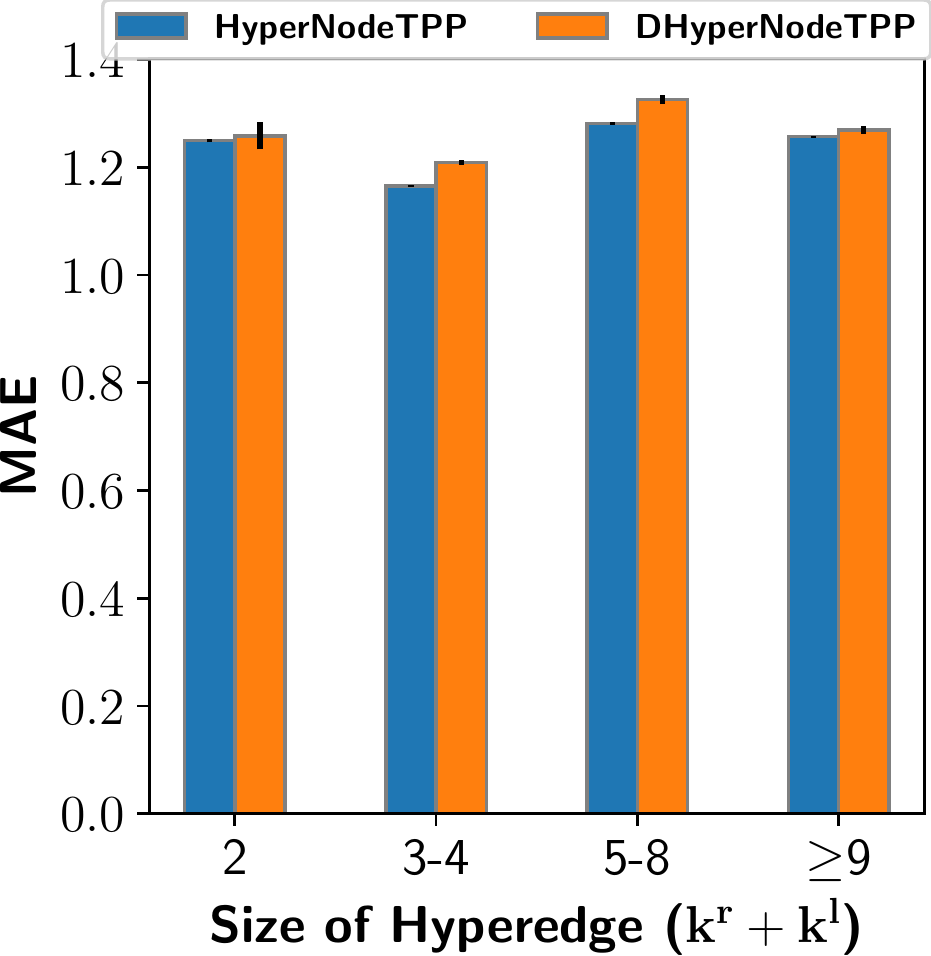}
    \caption{ Event time prediction}
    \label{fig:hepth_mae}
  \end{subfigure}
  \caption{Comparison of the performance of our directed and undirected model on different forecasting tasks on the Hepth dataset. Here, we can observe that representation from DHyperNodeTPP performs better than HyperNodeTPP for adjacency forecasting. Furthermore, DhyperNodeTPP performs considerably better than HyperNodeTPP for the event type prediction. For event time prediction, both models perform equally, as events are modeled on nodes, and for hyperedge size prediction, directed models perform better. Hence, we can learn better representation using direction information. }
  \label{fig:multi_task_hepth_performance}
\end{figure*}

\begin{figure*}
  \centering
  \begin{subfigure}{0.31\textwidth}
    \includegraphics[width=0.95\textwidth]{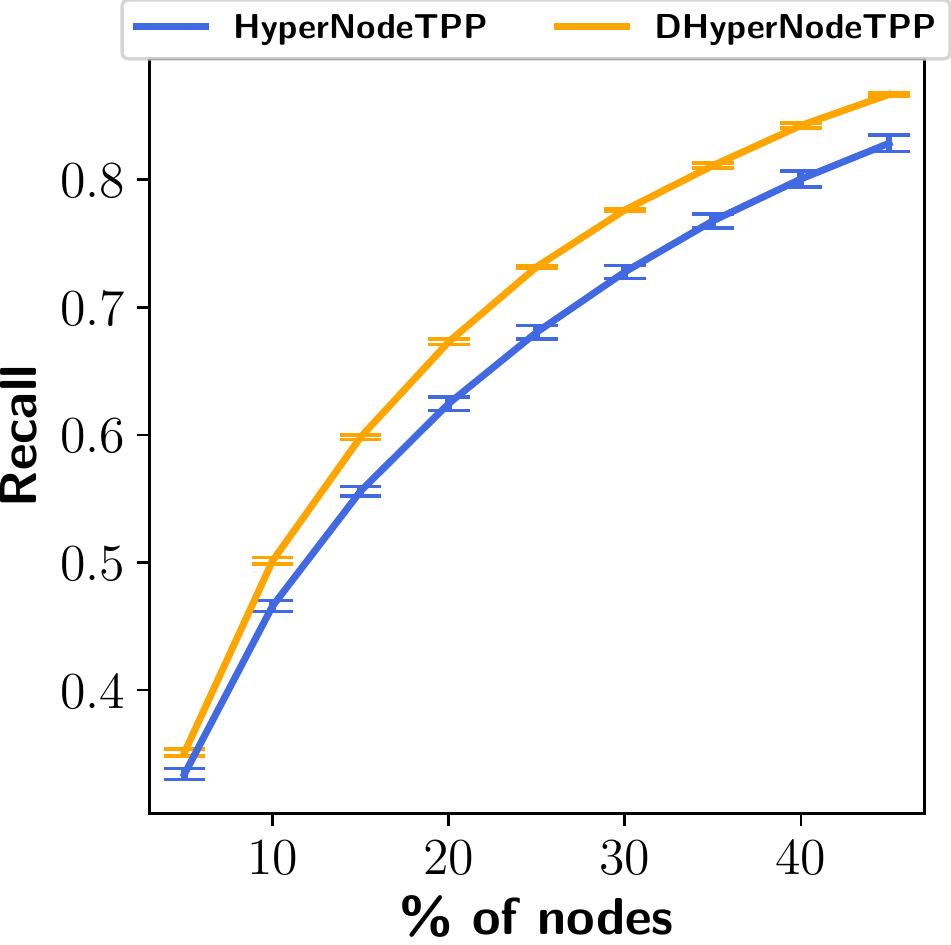}
    \caption{Adjacency forecasting}
    \label{fig:arxiv_connectivity}
  \end{subfigure}
  \begin{subfigure}{0.31\textwidth}
    \includegraphics[width=0.95\textwidth]{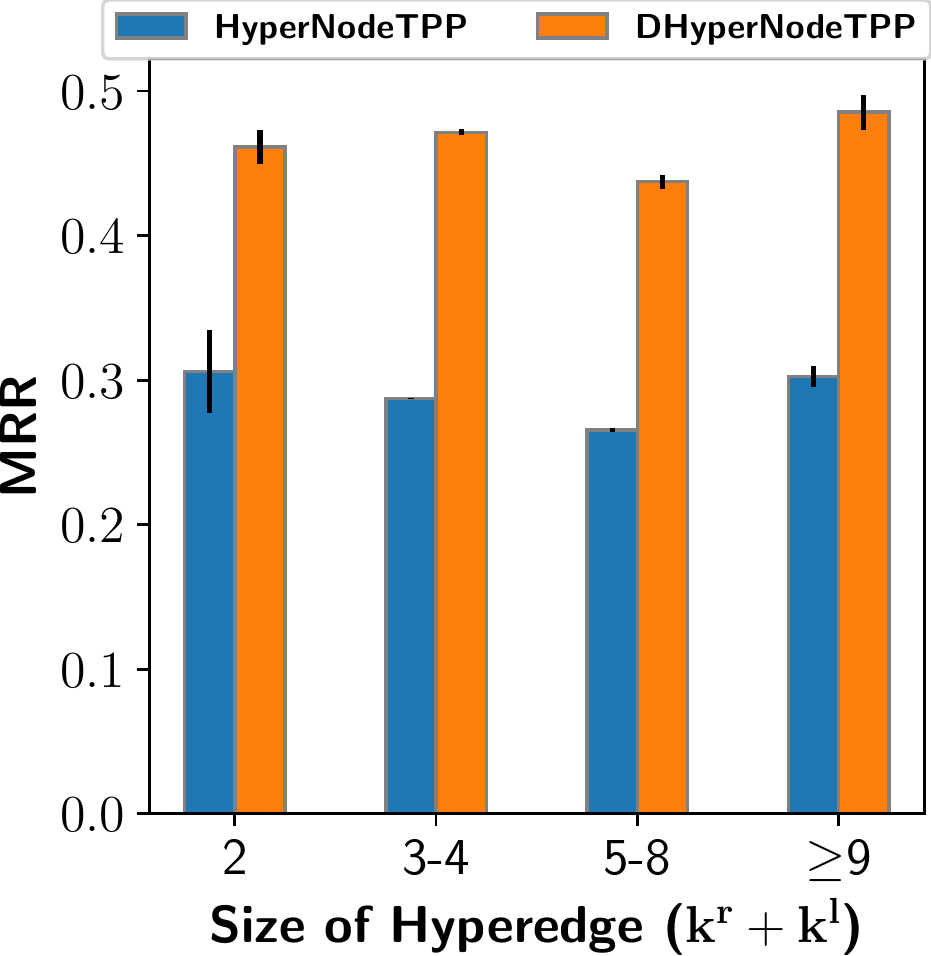}
    \caption{ Event type prediction}
    \label{fig:arxiv_mrr}
  \end{subfigure}
  \begin{subfigure}{0.31\textwidth}
    \includegraphics[width=0.95\textwidth]{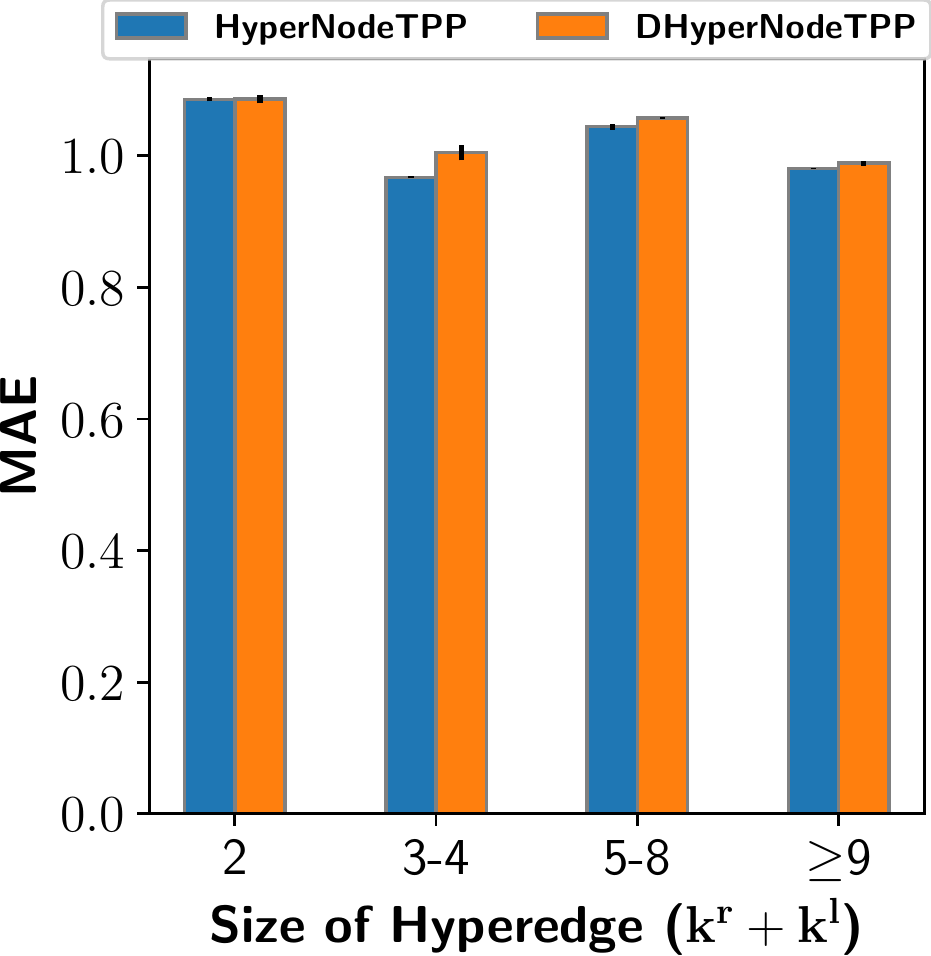}
    \caption{ Event time prediction}
    \label{fig:arxiv_mae}
  \end{subfigure}
  \caption{Comparison of the performance of our directed and undirected model on different forecasting tasks on ML Arxiv dataset. Here, we can observe that representation from DHyperNodeTPP performs better than HyperNodeTPP for adjacency forecasting. Furthermore, DhyperNodeTPP performs considerably better than HyperNodeTPP for the event type prediction. For event time prediction, both models perform equally, as events are modeled on nodes, and for hyperedge size prediction, directed models perform better. Hence, we can learn better representation using direction information. }
  \label{fig:multi_task_arxiv_performance}
\end{figure*}

\begin{figure*}
  \centering
  \begin{subfigure}{0.31\textwidth}
    \includegraphics[width=0.95\textwidth]{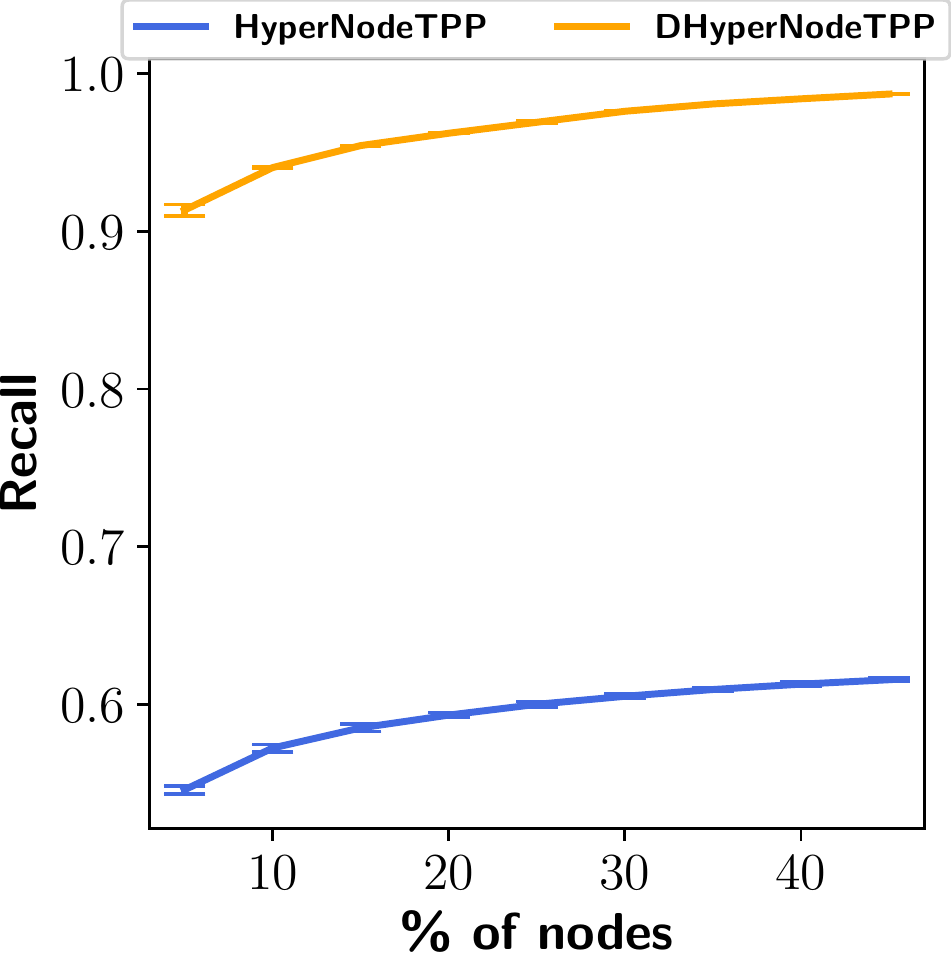}
    \caption{Adjacency forecasting}
    \label{fig:bitcoin_connectivity}
  \end{subfigure}
  \begin{subfigure}{0.31\textwidth}
    \includegraphics[width=0.95\textwidth]{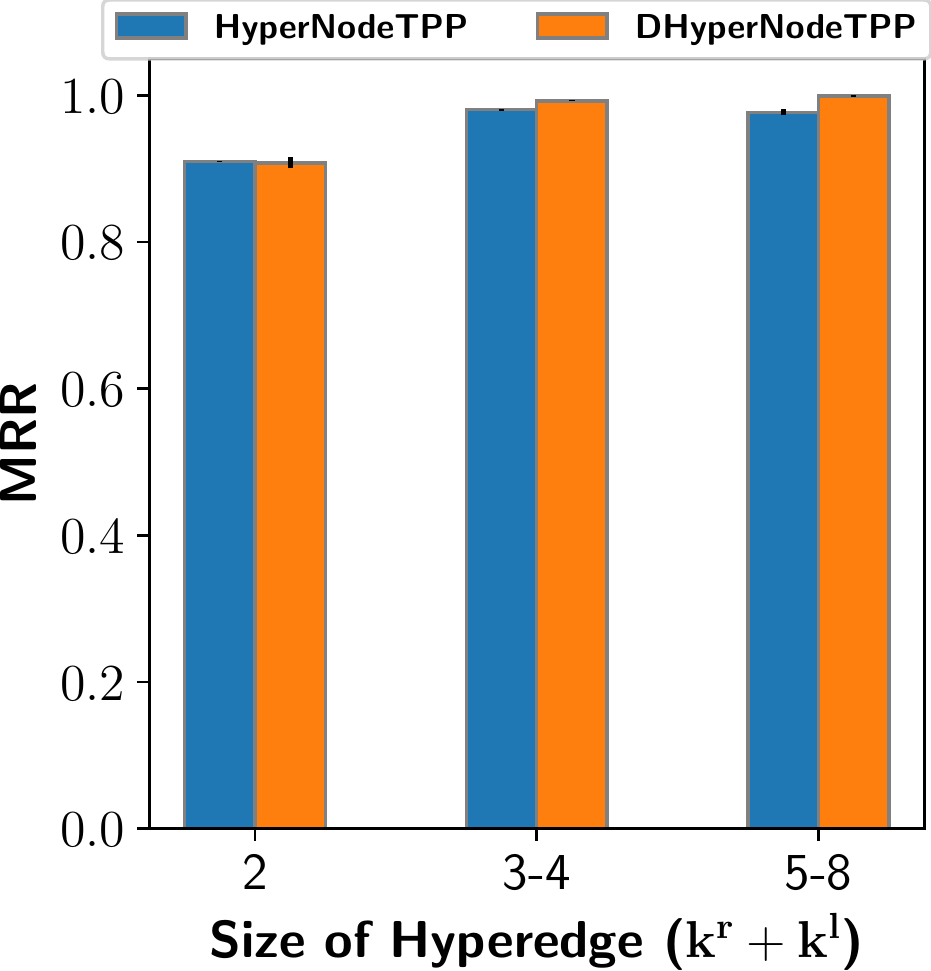}
    \caption{ Event type prediction}
    \label{fig:bitcoin_mrr}
  \end{subfigure}
  \begin{subfigure}{0.31\textwidth}
    \includegraphics[width=0.95\textwidth]{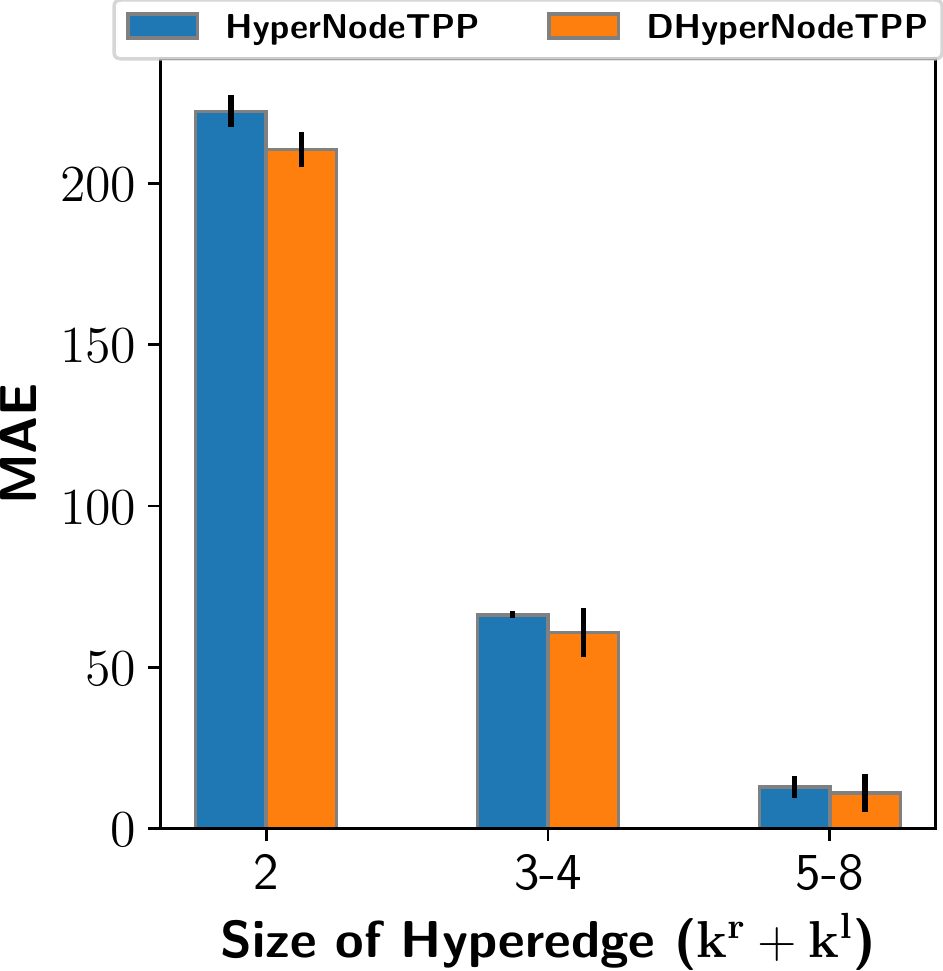}
    \caption{ Event time prediction}
    \label{fig:bitcoin_mae}
  \end{subfigure}
  \caption{Comparison of the performance of our directed and undirected model on different forecasting tasks on Bitcoin dataset. Here, we can observe that representation from DHyperNodeTPP performs better than HyperNodeTPP for adjacency forecasting. Furthermore, DhyperNodeTPP performs slightly better than HyperNodeTPP for the event type prediction. For event time prediction, both models perform equally, as events are modeled on nodes, and for hyperedge size prediction, directed models perform better. Hence, we can learn better representation using direction information. }
  \label{fig:multi_task_bitcoin_performance}
\end{figure*}

To estimate the effect of hyperedge size on performance, we grouped hyperedges based on the total hyperedge size ($k=k^r +k^\ell)$ and compared the performance of event type and time prediction tasks for DHyperNodeTPP and HyperNodeTPP. Here, hyperedge groups have the following sizes $k=2$ , $3 \leq k \leq 4$, $5 \leq k \leq 8$, and $\leq 9$. Figures \ref{fig:enron_mrr}, \ref{fig:eumail_mrr}, \ref{fig:twitter_mrr}, \ref{fig:hepth_mrr}, and \ref{fig:arxiv_mrr} show the comparison of MRR metrics for event type prediction tasks for DHyperNodeTPP and HyperNodeTPP models. For Enron-email, Eu-Email, and Twitter datasets, we can observe that HyperNodeTPP performs better than DHyperNodeTPP for hyperedge of size two. For hyperedges of size greater than two, DHyperNodeTPP outperforms HyperNodeTPP. In these datasets, Enron-Email, Eu-Email, and Twitter have around $80\%$, $81\%$, and $68\%$ hyperedges of size two ($k=2$), respectively, and HyperNodeTPP over-fits on these hyperedges which results in HyperNodeTPP outperforming DHyperNodeTPP when overall MRR is calculated, as shown in Table \ref{tab:dynamicdirectedlinkprediction}. For Hepth and ML-Arxiv datasets, DHyperNodeTPP outperforms HyperNodeTPP in all the hyperedge groups. Both models perform equally in all the hyperedge groups for the event time prediction, as shown in Figures \ref{fig:enron_mae}, \ref{fig:eumail_mae}, \ref{fig:twitter_mae}, \ref{fig:hepth_mae}, \ref{fig:arxiv_mae} and \ref{fig:bitcoin_mae}. This is because event times are estimated at the node level, and hyperedge size does not affect performance.

\subsection{Size prediction}
\label{appendix:size_prediction}
The performance of our model on the size prediction task in the candidate generation module is compared against the undirected baseline model. We evaluated the performance using AUC-Micro and AUC-Marco scores as shown in Table \ref{tab:size_prediction_auc_metrics}.  Figures \ref{fig:size_macro} and \ref{fig:size_prediction_auc_metrics_micro} visualized the same scores in a bar plot. Here, we can see our model DHyperNodeTPP outperforms HyperNodeTPP considerably. There is an average improvement of $13\%$ and $8\%$ in AUC macro and micro metrics, respectively.

\subsection{Adjacency vector prediction}
\label{appendix:adjacency_vector_prediction}
Figures \ref{fig:enron_connectivity}, \ref{fig:eumail_connectivity}, \ref{fig:twitter_connectivity}, \ref{fig:hepth_connectivity},  \ref{fig:arxiv_connectivity}, and \ref{fig:bitcoin_connectivity} show the comparison of DHyperNodeTPP and HyperNodeTPP for the adjacency vector prediction task in the candidate generation module using the recall metric. It is calculated at different thresholds corresponding to the percentage of nodes in the estimated adjacency vectors. Here, we can see that our model DHyperNodeTPP outperforms HyperNodeTPP in all datasets except in Eu-Email dataset. The improvement is huge for datasets Hepth, ML-Arxiv, and Bitcoin. Hence, we can conclude that the directed hyperedge assumption leads to better performing models.

\section{Additional Experiments} \label{sec:appendix:additional_experimentes}

\subsection{Scalable Training.}
\label{sec:appendix:scalable_training}
Figure \ref{fig:gain_in_speed} shows the gain in training speed obtained when we increased the batch size to 128 from 32. This is the ratio of time it took for DHyperNodeTPP to do a complete iteration of the dataset with batch size 32 divided by the time it took with batch size 128. Ideally, increasing batch size should reduce the computational time as it reduces the number of update steps and does more parallel computing within the GPUs, and we can observe the reduction in computation time for datasets Enron-Email, Eu-Email, Hepth, ML-Arxiv, and Twitter datasets. However, for the Bitcoin dataset, an increase in computation time was observed, with the time saturating around two hours across batch sizes of 32, 64, 128, and 256. This is likely due to GPU saturation and I/O bottlenecks, as Bitcoin is the largest dataset used in our study. Further, gains in training speed can be achieved by utilizing parallel computation techniques using multiple GPUs and CPUs. Recent work by \citet{ZhouEtAL:TGLAGeneralFrameworkForTemporalGNNTrainingOnBillionScaleGraphs} explores data structures that allow parallel GPU computation on the training samples.

\subsection{Comparing with state-of-the-art event prediction model} \label{sec:appendix:comparing_with_state_of_the_art_event_prediction_model}
In Table \ref{tab:eventprediction}, we compare the performance of our models, DHyperNodeTPP and HyperNodeTPP with the state-of-the-art event prediction model THP \citep{ZuoEtAL:2020:TransformerHawkesProcess}. 
Here, we observe a reduction in the MAE metric by $26\%$ for both models compared to THP. This improvement is due to our models' use of temporal representation of nodes to parameterize the node event model, incorporating all historical aggregated data. In contrast, the THP model's history is truncated based on the context length, leading to a loss of information. This performance reduction is particularly severe in datasets with a large number of events, as observed in the Bitcoin dataset.

\section{Limitations}
\label{sec:limiations}
For real-world forecasting, the proposed model requires temporal node representations for every node in the network $\bfV (t) \in \bbR^{|\calV| \times d}$. The node event model utilizes the representations to predict the timing of events for each node in the network, and the representations need to be computed at each event time. 
This is a computationally expensive task as the architecture mentioned in Section \ref{sec:dynamic_node_embedding} requires $\calO ( |\calV| \calN )$ computations. Hence, the time complexity of these computations can affect the model's application on domains where higher frequency events occur. 
Further, the model requires storing each node's recent $\calN$ relations to calculate its neighborhood features as explained in Section \ref{sec:dynamic_node_embedding}.
These stored relations are updated for the respective node at each time when an event involving the node has occurred. Hence, choosing hyperparameter $\calN$ requires considering both space and time complexity. These limitations mentioned here apply to the proposed model and the state-of-the-art pairwise edge forecasting model TGN~\citep{RossiEtAL:2020:TemporalGraphNetworksForDeepLearningOnDynamicGraphs}. 
Also, the model's performance in the candidate generation module needs to be considerably improved to reduce the search space and improve the accuracy of forecasting event type, as the recall in the adjacency vector forecasting task is low in the predicted most probable nodes, as shown in  Figures \ref{fig:enron_connectivity}, \ref{fig:eumail_connectivity}, \ref{fig:twitter_connectivity}, \ref{fig:hepth_connectivity}, \ref{fig:arxiv_connectivity}, and \ref{fig:bitcoin_connectivity}.

\begin{table}
\centering
\small 
\setlength{\tabcolsep}{1.2pt}
\begin{tabular}{lccc}
\toprule
\textbf{Methods}  & \textbf{THP}  & \textbf{HyperNodeTPP} & \textbf{DHyperNodeTPP}  \\ 
\midrule
\textbf{Enron-Email}  & 7.66 $\pm$ 1.54  & $\bm{4.15 \pm 0.01}$ & 4.18 $\pm$ 0.02  \\
\textbf{Eu-Email}   & 12.90 $\pm$ 0.13 & 12.23 $\pm$ 0.03 & $\bm{12.22 \pm 0.02}$ \\
\midrule
\textbf{Twitter} &  1.30 $\pm$ 0.06 & $\bm{1.18 \pm 0.01}$ & 1.24 $\pm$ 0.00 \\
\midrule 
\textbf{Hepth}  &  1.71 $\pm$ 0.20 & 1.25 $\pm$ 0.03 & $\bm{1.20 \pm 0.01}$ \\
\textbf{ML-Arxiv}  & 2.10 $\pm$ 0.95 & 1.25 $\pm$ 0.00 & $\bm{1.24 \pm 0.00}$ \\
\midrule 
\textbf{Bitcoin}   &  109.67 $\pm$ 2.84 & 75.98 $\pm$ 1.44 & $\bm{75.73 \pm 2.60}$ \\
\bottomrule
\end{tabular}
\caption{Performance of our Node Event Model in Section \ref{sec:model} compared to THP \citep{ZuoEtAL:2020:TransformerHawkesProcess}  state-of-the-art event prediction. Here, the MAE metric is used to evaluate the performance. We can observe that our proposed event prediction model based on network representations performs better than THP.}
\label{tab:eventprediction}
\end{table}
\end{document}